\documentclass[a4paper,fleqn]{cas-dc}

\usepackage[numbers]{natbib}
\usepackage{makecell}

\begin{document}
\shortauthors{Z. Zhang et~al.}
\shorttitle{DyArtbank}
\title [mode = title]{DyArtbank: Diverse Artistic Style Transfer via Pre-trained Stable
	Diffusion and Dynamic Style Prompt Artbank}

\author[1]{Zhanjie Zhang}[orcid=0000-0002-8966-1328]
\author[1]{Quanwei Zhang}
\author[1]{Guangyuan Li}
\author[1]{Junsheng Luan}   
\author[1]{Mengyuan Yang}
\author[2]{Yun Wang}
\author[1]{Lei Zhao\corref{cor1}}
                
\affiliation[1]{College of Computer Science and Technology, Zhejiang University, No. 38, Zheda Road, Hangzhou 310000, China}
\affiliation[2]{Department of Computer Science, City University of Hong Kong, Tat Chee Avenue, Kowloon, Hong Kong SAR}

\cortext[cor1]{Corresponding author}
\nonumnote{E-mail addresses: cszzj@zju.edu.cn (Z. Zhang), cszqw@zju.edu.cn (Q. Zhang), cslgy@zju.edu.cn (G. Li), l.junsheng121@zju.edu.cn (J. Luan), yangmy412@zju.edu.cn (M. Yang), ywang3875-c@my.cityu.edu.hk (Y.Wang), cszhl@zju.edu.cn (L. Zhao)}

\begin{abstract}
	Artistic style transfer aims to transfer the learned style onto an arbitrary
content image. However, most existing style transfer methods
can only render consistent artistic stylized images, making it difficult
for users to get enough stylized images to enjoy. To solve this
issue, we propose a novel artistic style transfer framework called
DyArtbank, which can generate diverse and highly realistic artistic
stylized images. Specifically, we introduce a Dynamic Style Prompt
ArtBank (DSPA), a set of learnable parameters. It can learn and store
the style information from the collection of artworks, dynamically
guiding pre-trained stable diffusion to generate diverse and highly
realistic artistic stylized images. DSPA can also generate random artistic image samples
with the learned style information, providing
a new idea for data augmentation. Besides, a Key Content Feature Prompt (KCFP) module is proposed to provide sufficient content prompts for pre-trained stable diffusion to preserve the detailed structure of the input content image. Extensive
qualitative and quantitative experiments verify the effectiveness of
our proposed method. Code is available: https://github.com/Jamie-Cheung/DyArtbank
\end{abstract}


\maketitle

\section{Introduction}
Artistic style transfer involves transferring the learned style onto
an arbitrary content image while preserving the detailed structure of the content image. The existing artistic style transfer methods generally can be divided into two categories: consistent artistic style transfer (CAST) methods and diverse artistic style transfer
(DAST) methods.

\begin{figure*}[htb]	
	\centering
	%	\vspace{-4.5cm}
	\includegraphics[width=1\textwidth,height=0.28\textwidth]{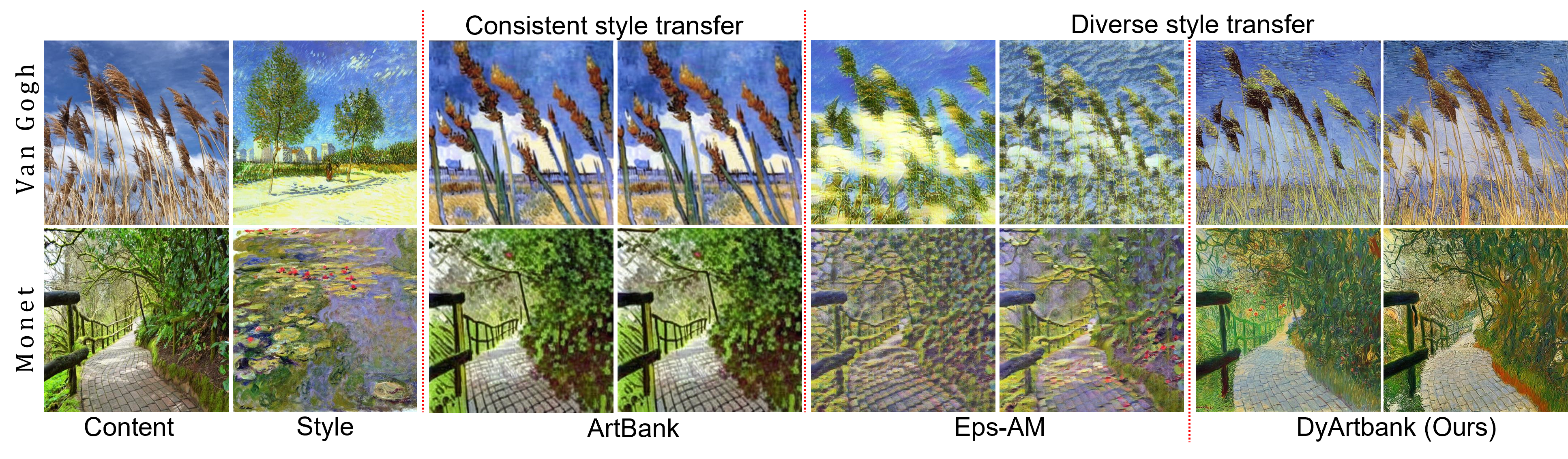} 
	\caption{Artistic stylized examples. The $1^{st}$ and $2^{nd}$ columns show the input content and style images. The $3^{rd}$ and $4^{th}$ columns show the stylized images synthesized by consistent style transfer methods (e.g., ArtBank ~\cite{zhang2024artbank}). Despite performing multiple inferences, ArtBank only obtained consistent artistic stylized images (i.e., the same content structure and style appearance). The other columns show the stylized images with the same content structure and different style appearance generated by diverse style transfer methods (e.g., Eps-AM~\cite{cheng2023user}, our proposed DyArtbank). \textbf{Note}: the random seeds are fixed for all methods.}
	\setlength{\belowcaptionskip}{-200cm}  
	\label{image2}
\end{figure*}

More specifically, consistent artistic style transfer methods~\cite{chen2023testnerf,zhang2021generating,zhang2024artbank,zhu2017unpaired,he2024progressive,liu2024intrinsic,qu2024source,li2023soft,chen2024towards,zhang2023caster}
generally refer to transferring the learned style information onto an arbitrary content image, rendering the consistent
artistic stylized image. Given an input content image, we perform multiple inferences using the CAST method and only get multiple consistent artistic stylized images with the same content structure and style appearance. For example, Zhu et al.~\cite{zhu2017unpaired}
introduced a CAST method that utilized generative adversarial network
and cycle consistency loss to learn the mapping between an
input content image and a style image. Zhang et al.~\cite{zhang2024artbank} proposed
a global style prompt to learn the style information from the
collection of artworks and condition large-scale pre-trained stable
diffusion model to generate highly realistic artistic stylized images.
However, although these methods can produce some consistent
artistic stylized images, they have difficulty in generating
diverse artistic stylized images (e.g., As shown in the $3^{rd}$ and $4^{th}$
columns of Fig.~\ref{image2}, ArtBank can only generate some artistic stylized
images with the same content structure and style appearance).

In contrast, given an arbitrary content image, diverse artistic
style transfer methods~\cite{chen2021diverse,cheng2023user,chu2024attack,wang2020diversified,huang2018multimodal,yang2022gating} can render diverse artistic
stylized images with the same content structure and different style
appearance. For example, Chu et al.~\cite{wang2020diversified} proposed to attack the
pre-trained deterministic generative models in order to achieve
diverse artistic stylized images. Cheng et al.~\cite{cheng2023user} proposed to utilize
the Sinkhorn-Knopp algorithm~\cite{cuturi2013sinkhorn} as entropy regularization to
perform diverse artistic style transfer. Wang et al.~\cite{wang2020diversified} proposed introducing deep feature disturbance during the stylization process to achieve diverse stylized images.
However, although these
methods can synthesize diverse artistic stylized images, they fail in
generating highly realistic artistic stylized images (e.g., As shown in
the $5^{th}$ and $6^{th}$ columns of Fig.~\ref{image2}, Eps-AM always introduce
some obvious artifacts and disharmonious patterns into the artistic
stylized images).

To synthesize diverse and highly realistic artistic stylized images,
we propose a novel method called DyArtbank, which can drag out
the massive prior knowledge from the large-scale pre-trained stable
diffusion and condition it to generate artistic stylized images. Specifically, we first introduce a novel Dynamic Style Prompt Artbank (DSPA), a
set of learnable parameters, which can learn the style information
from the collection of artworks and dynamically condition pretrained
stable diffusion to generate diverse and highly realistic
artistic stylized images. These learnable parameters can be treated
as a distribution that describes the visual attributes of the artwork collection (e.g., foreground, style, brush stroke, texture, etc.).  
A new parameter can be sampled from this distribution to guide pre-trained
SD to generate diverse artistic stylized images. Besides, we can
also utilize DSPA to randomly generate some artistic style image
samples, which provides a new idea for data augmentation. Finally,
we propose a novel Key Content Feature Prompt (KCFP) module,
which can guide the pre-trained stable diffusion to preserve the
content structure of the input content image. To summarize, our
contributions are listed as follows:

\begin{itemize}
	\item We introduce a novel Dynamic Style Prompt Artbank (DSPA),
	which can learn the style information from the collection
	of artworks and dynamically condition pre-trained stable
	diffusion to generate diverse artistic stylized images. Note that DSPA can also support the generation of some artistic style image samples, which opens a new idea for data augmentation.
	\item We propose a novel Key Content Feature Prompt (KCFP)
	module, which can provide sufficient content prompts for
	pre-trained stable diffusion to preserve the content structure
	of the input content image.
	\item To our knowledge, our proposed DyArtbank is the first to
	use pre-trained stable diffusion to accomplish diverse style
	transfer. Extensive qualitative and quantitive experiments
	demonstrate that our proposed DyArtbank achieves diverse
	and highly realistic artistic stylized images compared with
	state-of-the-art methods.
\end{itemize}
This work is extended from the conference paper ``ArtBank: Artistic Style Transfer with Pre-trained Diffusion Model and Implicit Style Prompt Bank" which was published in AAAI 2024~\cite{zhang2024artbank}. In the preliminary version, given the fixed random seed, the ArtBank can only render consistent artistic stylized images. This is because only one learnable parameter was used in the past and it is not possible to sample new learnable parameters in each inference. To solve this issue, we propose a novel artistic style transfer framework called DyArtbank, which can generate diverse and highly realistic artistic stylized images. Specifically, we introduce a Dynamic Style Prompt ArtBank (DSPA), a set of learnable parameters. It can learn and store the style information from the collection of artworks. During inference, we begin by sampling random noise from a normal distribution. Next, we compute the mean and variance from the DSPA, and then we adjust the mean and variance of the sampled noise using the reparameterization trick~\cite{kingma2013auto}. This process allows us to obtain new parameters that align with the distribution defined by the learnable parameters in the DSPA. We use the newly obtained parameters to dynamically guide the pre-trained stable diffusion model, enabling it to generate diverse and highly realistic artistic stylized images. The DSPA module can also generate random artistic image samples with the learned style information, providing a new idea for data augmentation. Then, we use CycleGAN as baseline and build extensive experiments to demonstrate the effectiveness of data augmentation. Besides, we train a Key Content Feature Prompt (KCFP) module,
which provide sufficient content prompts for pre-trained stable
diffusion to preserve the content structure of the input content
image. This addresses the issue in previous methods where some stylized images often fail to maintain the structure of the content images. Finally, we perform quantitative and qualitative experiments to demonstrate that the proposed DyArtbank (current version) outperforms the ArtBank (preliminary version) in content preservation and style diversity. Besides, extensive ablation studies have verified the effectiveness of DyArtBank. 
\section{Related Work}
\subsection{Consistent Artistic Style Transfer.} Consistent artistic style transfer
(CAST) methods usually refer to transferring the learned style
information onto an arbitrary content image, rendering the consistent
artistic stylized images. Given an input content image, we perform multiple inferences using the CAST method and only get multiple consistent artistic stylized images with the same content structure and style appearance. For example, existing generative
adversarial network (GAN-based) artistic style transfer methods~\cite{park2020contrastive,zhu2017unpaired} proposed to learn style information from the collection
of artworks via playing a min-max game between content
images and style images. To improve the realism of the artistic stylized
images, some artistic style transfer methods introduced
contrastive learning~\cite{khosla2020supervised} to minimize further the content structure error
between content images and style images. Recently, the large-scale
pre-trained text-image model~\cite{podell2023sdxl,rombach2022high} opened up a new way for
generating highly realistic artistic stylized images. Zhang et al.~\cite{zhang2023inversion}
proposed to utilize an inversion-based style transfer method that
can efficiently learn the critical information from a style image and
transfer the learned style information onto an arbitrary content image.
Zhang et al.~\cite{zhang2023prospect} introduced an expanded textual conditioning
space that consists of a set of textual embedding, and each prompt
corresponds to a specific generation stage. Deng et al.~\cite{deng2022stytr2} utilized
a Transformer to fuse long-range and global style features
into content features, obtaining stylized images. Liu et al.~\cite{liu2024dual} propose a Dual-head Genre-instance Transformer framework to simultaneously capture the genre and
instance features for arbitrary style transfer.
Chung et al.~\cite{chung2024style} manipulated the features of self-attention layers~\cite{wang2024cost,wang2022spnet,liu2021pose,liu2021person} as the way the cross-attention mechanism~\cite{li2023rethinking,li2023self}
works.
While these artistic
style transfer methods can generate highly realistic artistic stylized
images, they struggle to generate diverse artistic stylized images.

\subsection{Diverse Artistic Style Transfer.} Diverse artistic style transfer
(DAST) can render diverse artistic stylized images. Given an arbitrary
content image, by performing multiple inferences, DAST
can only obtain diverse artistic stylized images (i.e., the same content
structure and different style appearance). For example, Cheng
et al.~\cite{cheng2023user} proposed to frame the diverse style transfer as optimal
transport, using the Sinkhorn-Knopp algorithm~\cite{cuturi2013sinkhorn} via entropy
regularization. Chu et al.~\cite{chu2024attack} proposed to attack the pre-trained
deterministic generative models by adding a micro perturbation
to the input condition, achieving diverse artistic stylized images.
Wang et al.~\cite{wang2020diversified} operated deep feature perturbation to perturb the
deep image feature maps. Chen et al.~\cite{chu2024attack} proposed to enforce an
invertible cross-space mapping, achieving diversity. Huang et al.~\cite{huang2018multimodal} reassemble the content code of the original image with a randomly selected style code from the target domain to achieve multi-style transfer. Although these
diverse artistic style transfer methods can synthesize diverse artistic
stylized images, they fail to generate highly realistic stylized
images.

Compared with consistent and diverse style transfer methods,
our proposed DyArtbank can generate diverse and highly realistic
artistic stylized images.

\begin{figure*}[htb]	
	\centering
	%	\vspace{-4.5cm}
	\includegraphics[width=1\textwidth,height=0.65\textwidth]{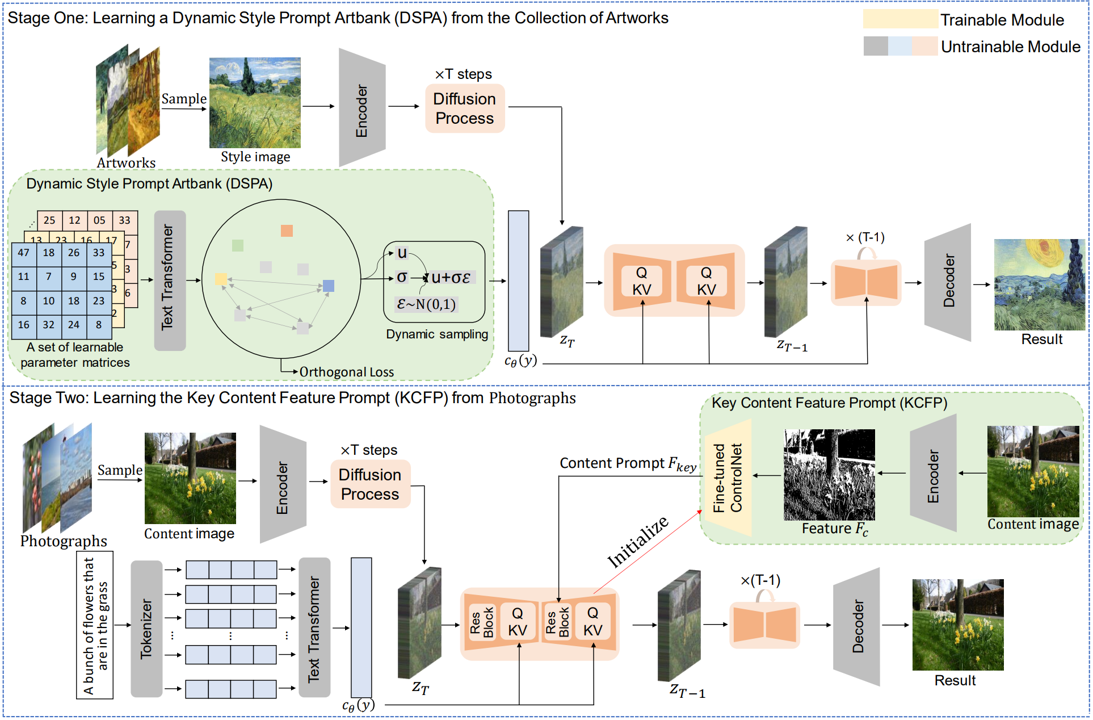} 
	\caption{The pipeline of training DyArtbank. In stage one, we learn a Dynamic Style Prompt Artbank (DSPA) to learn and store the style information from the collection of artworks. In stage two, we learn a Key Content Feature Prompt (KCFP) module to learn content prompts from the photographs.}
	\setlength{\belowcaptionskip}{-200cm}  
	\label{image3}
\end{figure*}

\begin{figure}[htb]	
	\centering
	%	\vspace{-4.5cm}
	\includegraphics[width=0.48\textwidth,height=0.25\textwidth]{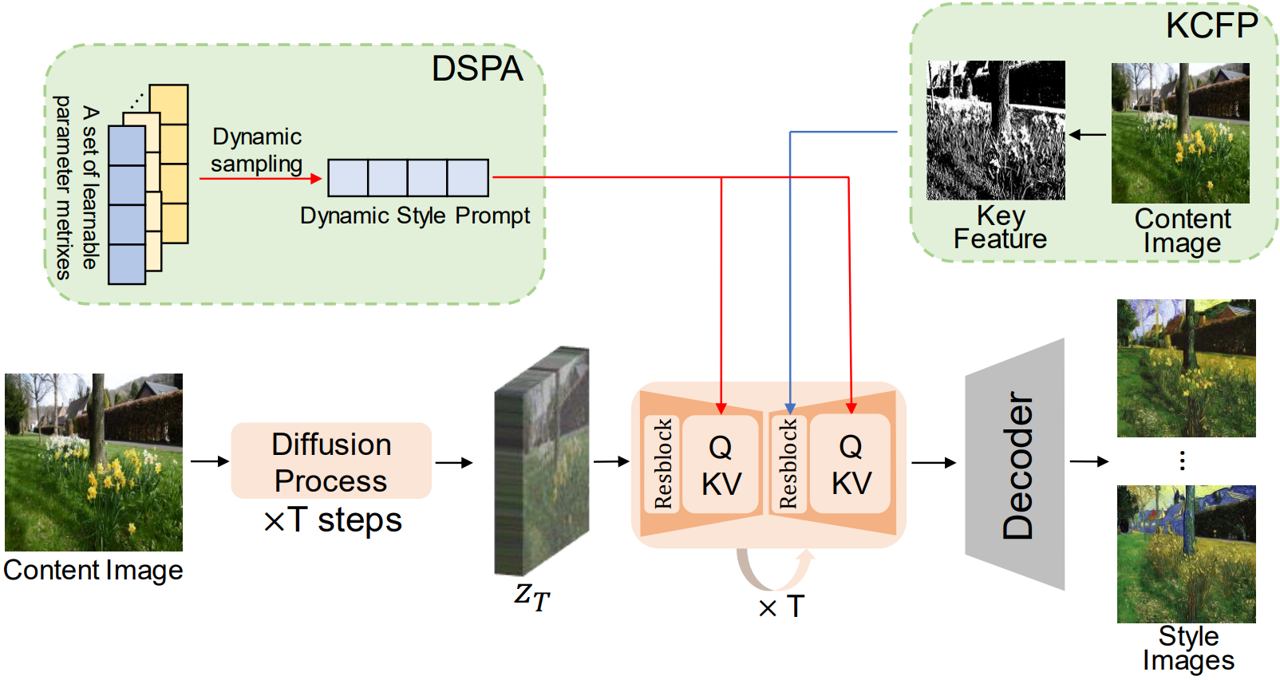} 
	\caption{The pipeline of artistic stylized image generation
		using DyArtbank. In inference,  a new parameter is sampled from DSPA,  which can dynamically guide the pre-trained stable diffusion to generate diverse artistic stylized images. Besides,
		to preserve the detailed structure of the input content image,
		the KCFP provides sufficient content prompts for pretrained
		stable diffusion.}
	\setlength{\belowcaptionskip}{-200cm}  
	\label{image4}
\end{figure}

\section{Method}
\subsection{Overview}
We aim to train a Dynamic Style Prompt Artbank (DSPA), a set of
learnable parameters which can learn the style information from the
collection of artworks and condition pre-trained stable diffusion
to generate diverse and highly realistic artistic stylized images.
Besides, we train a Key Content Feature Prompt (KCFP) module,
which provide sufficient content prompts for pre-trained stable
diffusion to preserve the content structure of the input content
image. As shown in Fig.~\ref{image3}, the pipeline of training our proposed
DyArtbank consists of two stages: learning a Dynamic Style Prompt
Artbank (DSPA) from the collection of artworks and learning the
Key Content Feature Prompt (KCFP) from photographs.

In stage one, we randomly sample style images from the collection
of artworks and denote a set of learnable parameters which learn
and store the style information from the collection of artworks.
Once the learnable parameters converge, we can dynamically sample
style prompts from them to condition pre-trained stable diffusion
to generate diverse and highly realistic artistic stylized images.

In stage two, we feed content images from photographs into
stable diffusion. Then, we fine-tuned a ControlNet which can extract
content prompt of the input content image to provide sufficient
content feature prompts for pre-trained stable diffusion to preserve
the content structure of the input content image.

The reason for training DSPA and KCFP separately is that DSPA is used to learn the unified style information from a collection of artworks, which often consists of hundreds of style images. This is insufficient for training KCFP; therefore, we train KCFP on content images~\cite{lin2014microsoft}.
As shown in Fig.~\ref{image4}, in inference, we can both utilize the
Dynamic Style Prompt Artbank (DSPA) and Key Content Feature
Prompt (KCFP) to condition pre-trained stable diffusion, thus generating
diverse and highly realistic artistic stylized images and preserving
the content structure of the input content image.

\subsection{Dynamic Style Prompt Artbank}
\label{DSPA}
Our goal is first to train a set of learnable parameters which can
learn and store style information from the collection of artworks.
Then  a new parameter is sampled dynamically from them to condition
large-scale pre-trained stable diffusion (SD)~\cite{rombach2022high,li2024rethinking} to generate diverse
and highly realistic artistic stylized images. Firstly, we argue that
SD utilizes the CLIP~\cite{radford2021learning} text encoder to condition the denoising
process, generating the desired image. Zhang et al.~\cite{zhang2024artbank} train a text
embedding $c$, a learnable parameter, to learn style information from
the collection of artworks using the following LDM loss~\cite{nichol2021improved}:

\begin{equation}
\mathcal{L}_{l d m}=\mathbb{E}_{z, x, c, t}\left[\left\|\epsilon-\epsilon_\theta\left(z_t, c, t\right)\right\|_2^2\right]
\end{equation}
where $z \sim E(x), \epsilon \sim \mathcal{N}(0,1)$. However, such a naive way fails to
generate diverse artistic stylized images.
%To generate diverse artistic stylized
%images, a naive way is to train multiple learnable parameters separately for a collection of artworks, making
%it possible to condition SD to obtain diverse artistic stylized
%images. However, training multiple trainable parameters requires extra
%computational burden and storage space. 
To this end, we propose a
Dynamic Style Prompt Artbank ($c^{*}$), which contains a set of learnable
parameters $\left(\mathcal{P}_1, \mathcal{P}_2, \ldots, \mathcal{P}_k\right)$ and condition SD to generate diverse
and highly realistic artistic stylized images. It is worth noting that
we train these learnable parameters simultaneously. Fewer learnable parameters can easily lead to the diversity degradation of stylized images or even DSPA training failure. Using more
learnable parameters will cost additional computation. Considering
extra computational burden, $k$ should satisfy $k \geq 4$, and $k=32$ is
the most reasonable.
Specifically,
we train a Dynamic Style Prompt Artbank using the following loss:
\begin{equation}
\mathcal{L}_{c^*}=\mathbb{E}_{z, x, c^*, t}\left[\left\|\epsilon-\epsilon_\theta\left(z_t, c^*, t\right)\right\|_2^2\right]
\end{equation}

where $c^* \in \mathcal{N}\left(\mu_{\mathrm{c}}, \sigma_{\mathrm{c}}^2\right)$. In fact, sampling $c^*$ from a stochastic distribution
is difficult for optimization. It is indispensable to utilize the
reparameterization trick that is used commonly in VAE~\cite{kingma2013auto}, then the optimization objective can be rewritten as below:

\begin{equation}
\mathcal{L}_{d s p a}=\mathbb{E}_{\mathrm{x}, \omega, \epsilon, t}\left[\left\|\epsilon-\epsilon_\theta\left(\mathrm{z}_t, \boldsymbol{\mu}_{\mathrm{c}}+\omega \sigma_{\mathrm{c}}, t\right)\right\|_2^2\right]
\end{equation}
where $\omega \sim \mathcal{N}(0,1)$. In inference, we randomly sample an $\omega$ and build reparameterization to guide pre-trained SD in generating diverse stylized images.  Besides, to further avoid a set of learnable
parameter matrices converge to the similar/same vector, which will
limit the generated artistic stylized image’s diversity. We further utilize
orthogonal loss~\cite{lu2022prompt} to penalize on the cosine similarity and encourage orthogonality between each learnable parameter matrix $\mathcal{P}$:

\begin{equation}
\mathcal{L}_{\text {ortho }}=\frac{1}{K(K-1)} \sum_{i=1}^K \sum_{j=i+1}^K\left|\left\langle\mathcal{E}\left(\mathcal{P}_i\right), \mathcal{E}\left(\mathcal{P}_j\right)\right\rangle\right|, K=32
\end{equation}
where $\mathcal{E}$ denotes the text encoder of CLIP~\cite{radford2021learning} and ⟨·, ·⟩ is cosine
similarity between each learnable parameter matrix. The final optimization
objective is shown below:
\begin{equation}
\mathcal{L}=\mathcal{L}_{d s p a}+\lambda \mathcal{L}_{\text {ortho }}
\end{equation}
where $\lambda$ is a hyperparameter (i.e., $\lambda$=5$\times$$10^{-3}$).

\subsection{Key Content Feature Prompt}
ControlNet~\cite{zhang2023adding} utilizes depth maps, canny maps, etc., to effectively
provide content prompts for pre-trained stable diffusion to preserve
the content structure. However, such content prompts are only effective
for text-image tasks, but they fail to provide enough content
prompts for artistic style transfer. Style transfer requires stylized
images to maintain the detailed structure of the input content image.
To this end, we introduce a novel Key Content Feature Prompt
(KCFP) module to further provide sufficient content prompts for
SD to preserve the detailed structure of the input content image. In
fact, our proposed KCFP consists of a fine-tuned ControlNet and a VAE
Encoder. We first utilize the VAE Encoder
$E$ of SD to extract the content feature $F_{c}$ . Then, we fine-tune
a ControlNet to extract the Key Content Feature Prompt (KCFP)
further, providing sufficient content prompts for pre-trained SD to
preserve the detailed structure of the input content image. Specifically,
we utilize the encoder $E$ of SD to obtain content prompts $F_{c}$
of input content image $I_{c}$ as below:
\begin{equation}
F_c=E\left(I_c\right), F_c \in R^{4 \times 64 \times 64}
\end{equation}
Then, we still copy the original weights of SD to initialize the weight
of ControlNet. The difference is that we use content prompts $F_{c}$
as an additional control to condition pre-trained SD to preserve the
content structure of input content images. Given a content image
$z_{0}$, we first utilize BLIP~\cite{li2022blip} to obtain its text prompt $c_{t}$ . Then,
we add noise to the image and produce a noisy image $z_{t}$, then we
fine-tune ControlNet using the following loss:
\begin{equation}
\left.\mathcal{L}_{k e y}=\mathbb{E}_{z_0, t, c_t, F_{\mathrm{c}}, \epsilon \sim \mathcal{N}(0,1)}\left[\| \epsilon-\epsilon_\theta\left(z_t, t, c_t, F_{\mathrm{c}}\right)\right) \|_2^2\right]
\end{equation}
After the training, the KCFP can provide sufficient content prompts
for pre-trained SD to preserve the content structure of the input content
image. To facilitate understanding, Table.~\ref{variables} summarizes the meanings of the variables used in this study.
\begin{table}[ht]
    \centering
    \caption{Summary of Variables}
    \begin{tabular}{@{}ll@{}}
        \toprule
        Variable & Description  \\ 
        $c$ & a learnable parameter  \\
        $c^{*}$ & a set of learnable
parameters $\left(\mathcal{P}_1, \mathcal{P}_2, \ldots, \mathcal{P}_k\right)$  \\
        $k$ & the number of the learnable
parameter \\
        $\mathcal{P}_k$& the k-th learnable parameter matrix  \\
        $\epsilon$& \makecell{a random noise that follows a standard \\ normal distribution $\mathcal{N}(0,1)$}\\
        $\epsilon_\theta$&\makecell{A parameterized model used for noise \\ prediction (Unet)}\\
        t &time step\\
        $z_{t}$&the feature representation at time step\\
        $\mathcal{E}$& text encoder of CLIP \\
        $\mu_{\mathrm{c}}$& the mean of the learnable parameter $c$\\
        $\sigma_{\mathrm{c}}$&the variance of the learnable parameter $c$\\
        $F_c$ & the content prompt of input image\\
        \bottomrule
    \end{tabular}
    \label{variables}
\end{table}

\begin{figure*}[htb]	
	\centering
	\includegraphics[width=1\textwidth,height=0.65\textwidth]{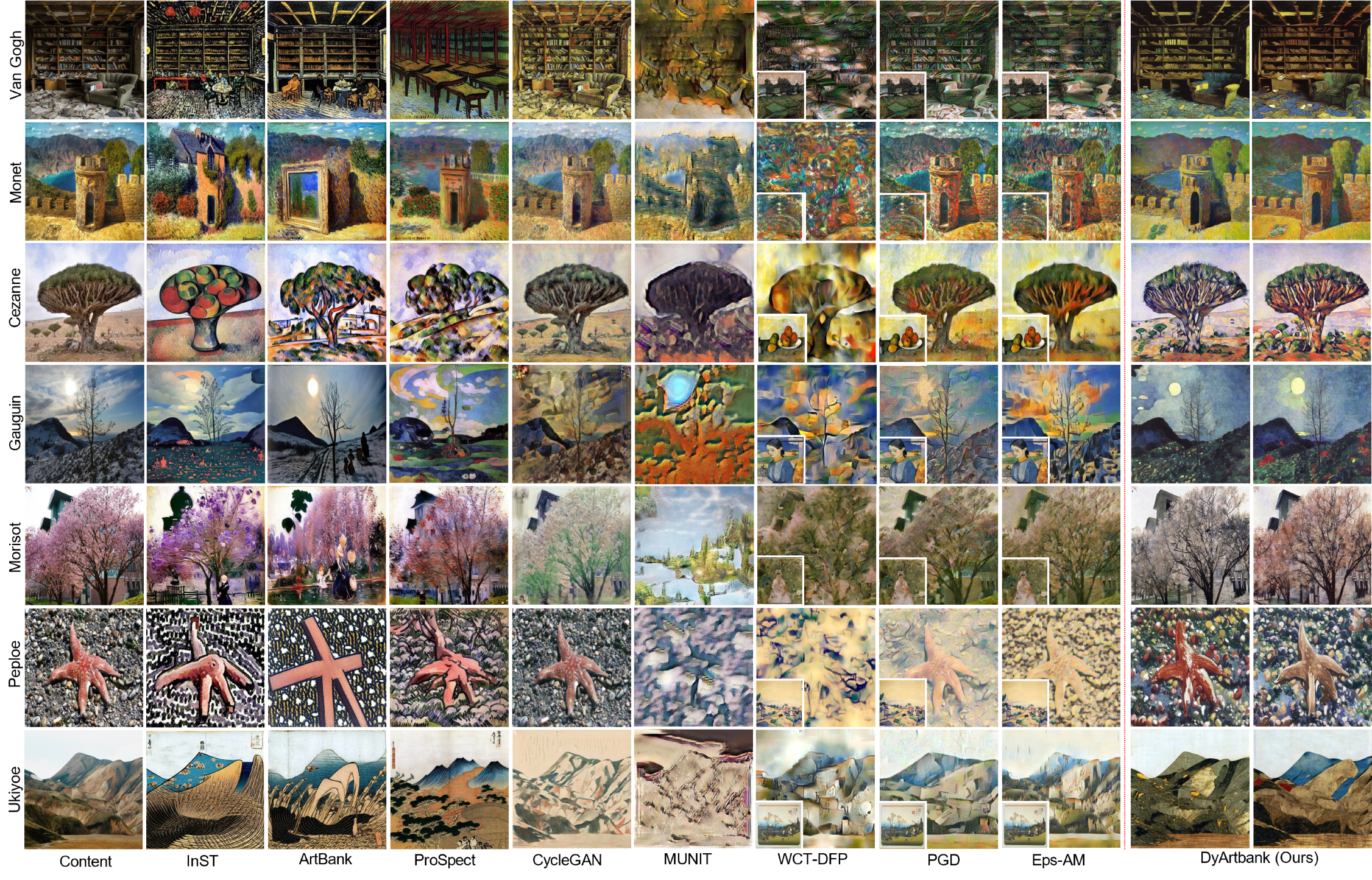} 
	\caption{Qualitative comparison. Compared to the previous version (Artbank)~\cite{zhang2024artbank}, the DyArtbank can preserve the content images' structure better.}
	\label{image5}
\end{figure*}

\begin{table*}[htb]
	\caption{The FID estimates the divergence between human-created artistic images and synthesized stylized images (lower is better). $^{*}$ denotes the average preference score.}
	\centering
	\setlength{\tabcolsep}{0.1cm}
        \small
	\begin{center}		
		\begin{tabular}{c|c|ccccccccc}
			\hline Metrics & Datasets & Ours & InST & ArtBank & LSAST & CycleGAN &MUNIT&WCT-DFP& PGD& Eps-AM \\
			\hline \multirow{7}{*}{ FID $\downarrow$} & Van Gogh &\textbf{105.36} & 113.26 & 121.62 & 109.34 & 112.04 &142.83&138.25& 135.27 & 137.94 \\
			& Morisot & \textbf{179.40} & 221.37 & 225.47 & 185.23 & 204.74 &251.69&249.27& 245.76 & 238.49 \\
			& Ukiyoe & \textbf{90.18} & 104.73 & 111.50 & 96.87 & 140.78 &167.25&157.94& 152.84 & 155.67 \\
			& Monet & \textbf{129.79} & 159.39 & 166.70 & 135.26 & 170.50 &192.34&158.72& 183.21 & 186.30 \\
			& Cezanne & \textbf{137.14} & 150.24 & 143.90 & 141.68 & 151.14 &168.52&165.33 & 160.25 & 163.71 \\
			& Gauguin & \textbf{122.58} & 145.65 & 155.27 & 127.49 & 160.00 &173.27&172.68& 171.18 & 169.42 \\
			& Peploe & \textbf{162.71} & 188.37 & 182.51 & 186.45 & 167.77 &199.74&194.17& 195.81 & 196.53 \\
			%		\hline \multirow{7}{*}{ CLIP $\uparrow$} & Van Gogh & 0.7426 & 0.7244 & 0.7321 & 0.7289 & 0.6875 & 0.6524 & 0.6631 \\
			%		 & Morisot & 0.7452 & 0.6983 & 0.7447 & 0.7235 & 0.7063 & 0.6728 & 0.6529 \\
			%		 & Ukiyoe & 0.7428 & 0.7272 & 0.7384 & 0.7401 & 0.6504 & 0.6307 & 0.6237 \\
			%		 & Monet & 0.7649 & 0.7319 & 0.7556 & 0.7328 & 0.7351 & 0.7071 & 0.7128 \\
			%		 & Cezanne & 0.7702 & 0.7332 & 0.7646 & 0.7682 & 0.7363 & 0.7102 & 0.7098 \\
			%		 & Gauguin & 0.8213 & 0.7875 & 0.8190 & 0.8021 & 0.7139 & 0.6921 & 0.6843 \\
			%		 & Peploe & 0.7423 & 0.7032 & 0.7355 & 0.7371 & 0.6846 & 0.6450 & 0.6537 \\
			
			\hline Time/sec $\downarrow$ & - & 4.2137 & 4.0485 & 3.7547 & 4.1325 & 0.0312 &0.0468&1.283& 0.7354 & 0.0720 \\
			Preference $\uparrow$ & - & $\textbf{0.67}^*$ & $0.55 / 0.45$ & $0.56 / 0.44$ & $0.59 / 0.41$ & $0.61 / 0.39$ &$0.82 / 0.18$&$0.80 / 0.20$& $0.72 / 0.23$ & $0.70 / 0.30$ \\
			Deception $\uparrow$ & - & \textbf{0.763} & 0.716 & 0.614 & 0.752 & 0.582 &0.428&0.439& 0.435 & 0.462 \\
			\hline
		\end{tabular}
	\end{center}
	\label{table1}
\end{table*}

%\begin{table*}[htb]
%	\caption{The LPIPS measures the content consistency between input content image and stylied image (lower is better).}
%	\centering
%	\setlength{\tabcolsep}{0.1cm}
%	\begin{center}		
%		\begin{tabular}{c|c|ccccccc}
%			\hline Metrics & Methods & Van Gogh & Morisot & Ukiyoe & Monet & Cezanne & Gauguin & Peploe \\
%			\hline \multirow{2}{*}{ LPIPS $\downarrow$} & w/ ControlNet & 0.6254 & 0.5127 & 0.5596 & 0.5056 & 0.6032 & 0.5284 & 0.5920 \\
%			& Ours & 0.4792 & 0.4971 & 0.4825 & 0.4718 & 0.5182 & 0.4934 & 0.5018 \\
%			\hline
%		\end{tabular}
%	\end{center}
%	\label{table2}
%\end{table*}

\begin{figure}[htb]	
	\centering
	\includegraphics[width=0.48\textwidth,height=0.23\textwidth]{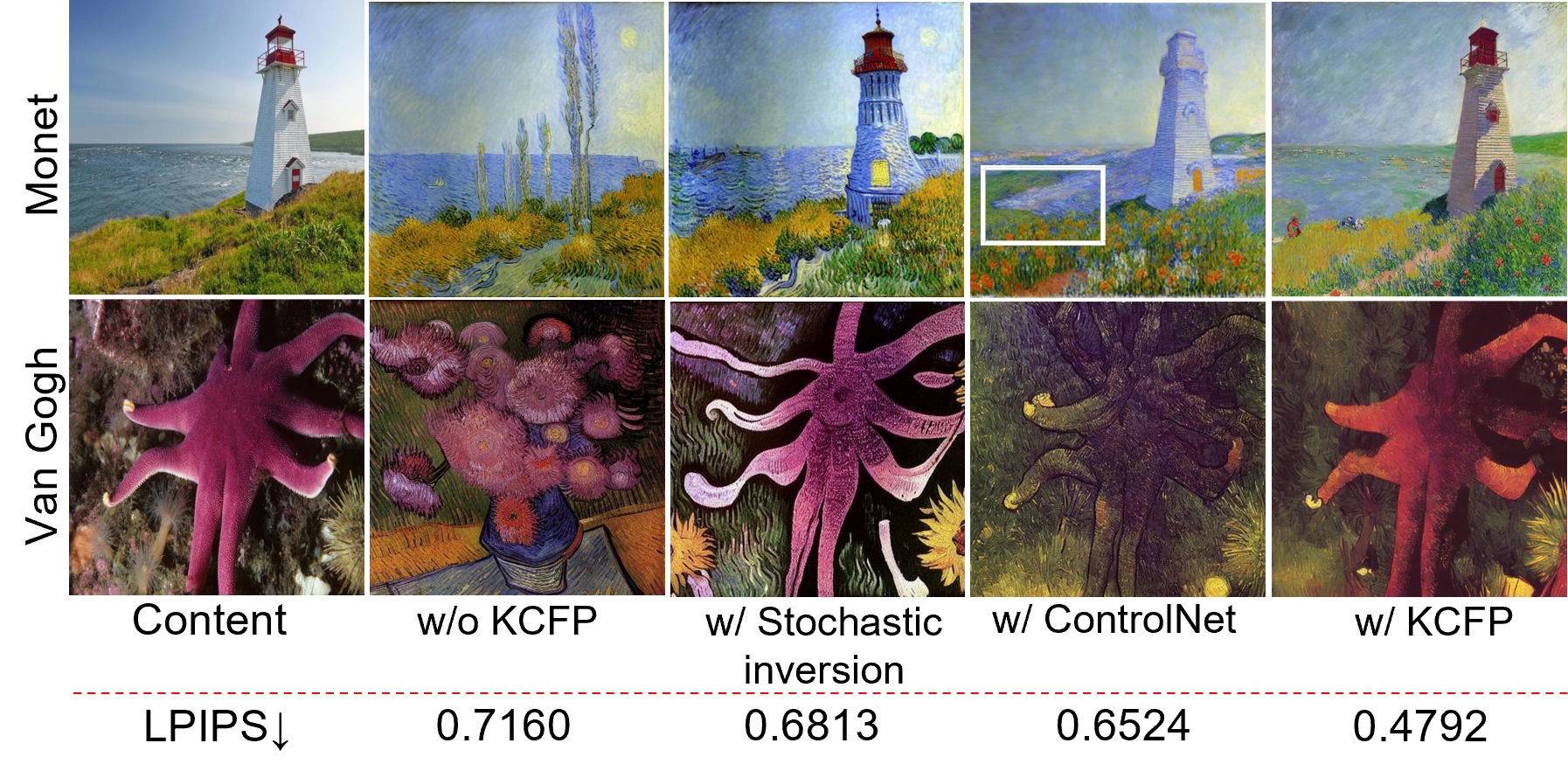} 
	\caption{Ablation study of KCFP.}
	\label{image7}
\end{figure}

%\begin{figure*}[htb]	
%	\centering
%	\includegraphics[width=1\textwidth,height=0.55\textwidth]{image8.pdf} 
%	\caption{Artistic image examples by randomly sampling from the Dynamic Styl Prompt Artbank. It is worth noting that we
%		do not provide additional text conditions or initialize content images as input.}
%	\label{image8}
%\end{figure*}

\subsection{The differences between ArtBank and DyArtbank}
As shown in Tab.~\ref{differences}, we summarize the main differences between Artbank and DyArtbank. Compared to ArtBank, DyArtbank utilizes 32 learnable parameters to learn the styles from the artists' collection instead of just one, and it supports the generation of diverse stylized images. Additionally, DyArtbank employs a two-stage training process, in which a Key Content Feature Prompt (KCFP) module is trained. In the inference, KCFP extracts the key content feature prompt to provide content prompts for stable diffusion to preserve the content structure of the input image, rather than relying solely on stochastic inversion~\cite{galimage} as ArtBank does.

\begin{table}[htb]
	\centering
    \caption{The main differences between Artbank and DyArtbank.}
	\begin{center}		
		\begin{tabular}{c|cc}
			\hline
			\ Types&ArtBank &DyArtbank
			\\
			\hline
			\ \makecell{The number \\ of  learnable \\ parameters}&1&32
			\\
                \hline
                \ \makecell{Content \\ Prompt}&None&\makecell{Content \\ Feature}
			\\
                \hline
			\ Diversity &\makecell{Not \\ Applicable}& Applicable
			\\
                \hline
			\ \makecell{Whether \\ multi-stage \\ training}  &No&Yes
			\\
			\hline 
		\end{tabular}
	\end{center}
	\label{differences}
\end{table}

\section{Experiments}
%This section provides details of implementing our DyArtbank and
%qualitative and quantitative comparisons to verify its superiority.
\subsection{Implementation Deatails}
We used pre-trained SD (version 1.5) as the backbone. The backbone
is not limited to SD (version 1.5) but includes other large-scale
pre-trained models such as SD (version 2.1)/ SD-XL. To train our
proposed Dynamic Style Prompt Artbank (DSPA), we collect nine artists’
artwork collections from WikiArt~\cite{nichol2016painter}. The size of each artwork
collection is 400 for Van Gogh, 1072 for Monet, 584 for Cezanne, 130
for Morisot, 204 for People, 1433 for Ukiyoe, 859 for Roeriuch, and 101
for Kandinsky. It is worth noting that each artwork collection needs to train a DSPA separately, sharing the same
KCFP.
We sample content images from COCO~\cite{lin2014microsoft} to train the Key Content Feature Prompt (KCFP) module, where  KCFP can extract content prompts from arbitrary content images and further condition pre-trained SD to preserve the content structure.

In training, all
artistic images and content images are scaled to 512×512 pixels,
and we set the learning rate as 0.0001. We build all the experiments
on two NVIDIA GeForce RTX3090 GPUs. In inference, we collect some high-quality content images
from DIV2K~\cite{agustsson2017ntire} as the initial input images. Besides, it is worth noting that the random seeds are fixed in all experiments.
\subsection{Qualitative comparisons}
As shown in Fig.~\ref{image5}, we compare our method with the state-of-the-art
consistent style transfer methods (e.g., InST~\cite{zhang2023inversion}, Artbank~\cite{zhang2024artbank}, LSAST~\cite{zhang2024towards}, CycleGAN~\cite{zhu2017unpaired}) and diverse style transfer methods
(e.g., Eps-AM~\cite{cheng2023user}, PGD~\cite{chu2024attack}, MUNIT~\cite{huang2018multimodal}, WCT-DFP~\cite{wang2020diversified}). As the representatives of the former,
InST, Artbank, and LSAST guide the large-scale pre-trained model
to generate artistic stylized images.
InST introduces some undesired content structure (e.g., $2^{nd}$ and $3^{rd}$
rows). Artbank struggles with preserving the detailed content structure (e.g., $2^{nd}$ and $5^{th}$
rows). LSAST introduces disharmonious patterns (e.g., $4^{th}$ rows). CycleGAN utilizes a generative adversarial
network to learn the mapping between content images and style
images, introducing some obvious artifacts (e.g., $5^{th}$ and $7^{th}$ rows).
As the representative of the latter, MUNIT always introduces obvious artifacts (e.g., $4^{th}$ rows) and sometimes fails to generate stylized images (e.g., $5^{th}$ rows). WCT-DFP always introduces excessive stylistic information which makes it difficult to maintain the structure of the content image. PGD and EpsAM train a well designed
forward network to render artistic stylized images. PGD
always introduces some disharmonious patterns (e.g., $2^{nd}$ and $7^{th}$
rows). Eps-AM produces undesired artifacts and smooth regions
(e.g., $3^{rd}$ and $4^{th}$ rows).

\subsection{Quantitative Comparisons.}
\textbf{FID}~\cite{heusel2017gans}. To effectively measure the image quality, we utilize the
widely-used FID (Fréchet Inception Distance), which can estimate
the distribution divergence between human-created style images
and generated stylized images in a deep feature space~\cite{szegedy2016rethinking}. To compute
FID, we take Van Gogh style as an example; for consistent
style transfer methods (i.e., InST, Artbank, LSAST, and CycleGAN), we randomly choose 6,000 input content images, achieving
6,000 stylized images to compute FID between 6,000 stylized images
and 400 style images from Van Gogh’s artworks. Similarly, for
diverse style transfer methods (i.e., PGD, Eps-AM), we randomly
select 150 content images and 400 style images, generating 6,000
stylized images to compute FID. As shown in Table.~\ref{table1}, our proposed
DyArtbank achieves a better FID score (lower is better).

%CLIP score [10]. CLIP (Contrastive Language-Image Pre-training)
%is a pre-trained model based on 400M text-image pairs and can measure
%the distance between text description and stylized images. Take
%Van Gogh style as an example; we use 6,000 content images for InST,
%Artbank, ProSpect, and CycleGAN to generate 6,000 stylized images.
%Then, we compute the CLIP score between the 6,000 stylized
%images and the text description (a painting by Van Gogh). Similarly,
%we randomly choose 150 content images and 400 style images for
%PDG and Eps-AM to generate 6,000 stylized images to calculate the
%CLIP score. As shown in Table. 1, our proposed method achieves a
%better CLIP score (higher is better).

\textbf{Timing Information}~\cite{huang2017arbitrary,sun20243dgstream,sun2023vgos}. The ``Time/sec" row of Tab.~\ref{table1}
shows the inference time comparison on images with a size of
512×512 pixels. The speed of our DyArtbank is only comparable
with the pre-trained stable diffusion-based methods (e.g., InST, Artbank,
LSAST) because we introduce an extra Key Content Feature
Prompt (KCFP). But there is no denying that our proposed method
can generate diverse and more highly realistic artistic stylized images.
Besides, DyArtbank can better preserve the input content
images’ content structure.

\textbf{Preference Score}~\cite{wang2022aesust,zhang2024rethink}. To verify the popularity of artistic
stylized images generated by our DyArtbank. We build an A/B user
study between our and the other style transfer methods. Specifically,
we randomly choose 100 content images for InST, Artbank, LSAST,
CycleGAN, and DyArtbank to generate 100 artistic stylized images.
Then, we randomly select 10 content images and 10 style images
for PGD and Epa-AM to generate 100 stylized images, respectively.
To ensure equal comparison, we asked each participant to select
their preferred stylized images between our method and one of
the other methods. We collect 2,000 votes from 200 participants for
each artistic style. The ``Preference" row of Table.~\ref{table1} shows that our
DyArtbank achieves a higher preference score, indicating that our
proposed method achieves more popularity.

\textbf{Deception Score}~\cite{sanakoyeu2018style}. To assess whether the generated images
by our DyArtbank are more likely to be distinguished as humancreated,
we build a user study to calculate the deception score. In this
user study, we randomly chose 50 generated artistic stylized images
for each method and asked subjects to guess whether they were
human-created images. The "Deception" row of Table.~\ref{table1} reports
the deception score (higher is better). Besides, we report that the
deception score from Wikiart~\cite{nichol2016painter} is 0.863. Our DyArtbank achieves
a higher deception score and is closer to the human-created artistic
image. Thus, our DyArtbank can generate more highly realistic
artistic stylized images.

\begin{figure*}[htb]	
	\centering
	\includegraphics[width=1\textwidth]{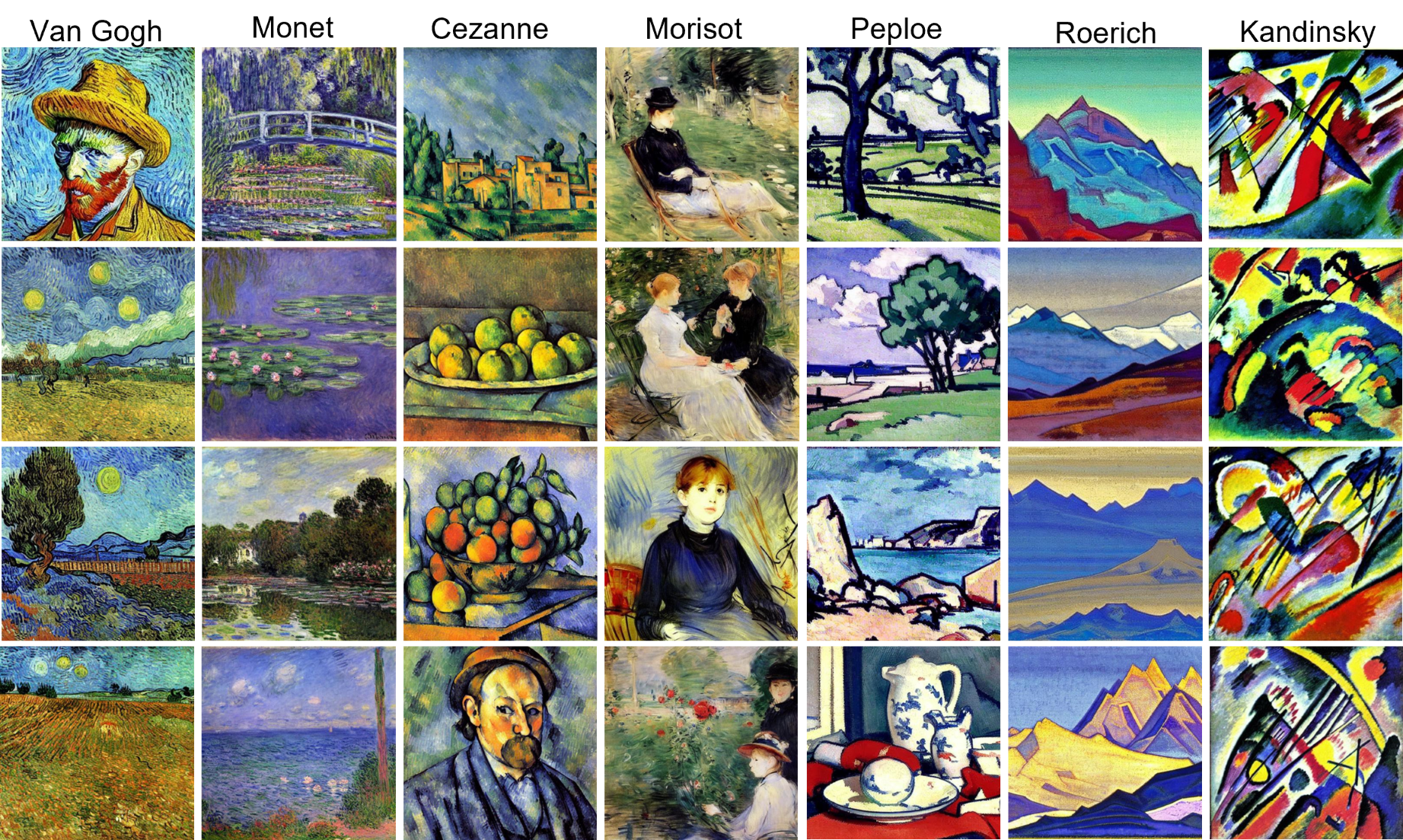} 
	\caption{Artistic image examples by randomly sampling from the Dynamic Styl Prompt Artbank. It is worth noting that we
		do not provide additional text conditions or initialize content images as input.}
	\label{image8}
\end{figure*}
\subsection{Ablation Study}
\textbf{Stochastic Sampling}. Our proposed DyArtbank can generate some artistic image examples (e.g., Van
Gogh, Monet, Cezanne, Morisot, Peploe, Roerich, and Kandinsky).
Take Van Gogh’s artworks; we feed style images from the collection
of Van Gogh’s artworks and train Dynamic Style Prompt Artbank
(DSPA) (as shown in Stage One of Fig.~\ref{image3}). Once DSPA converges,
we can use it to condition pre-trained stable diffusion to generate 
random artistic images using noise images. We show some artistic images generated by our method in Fig.~\ref{image8}. 

%\textbf{Data Augmentation}. Recently, large-scale pre-trained models~\cite{kang2023scaling,podell2023sdxl} were able to generate highly realistic images, mainly because
%they learn abundant knowledge from the massive data ~\cite{schuhmann2022laion}. It
%means that the amount of data is essential for training an effective
%model. Thus, we retrain the CycleGAN and CycleGAN-aug with the
%augmented data (i.e., The augmented data includes human-created
%style images and stable diffusion-generated style images). Specifically,
%for the Van Gogh style, we use 6,287 content images and 400
%Van Gogh style images to train CycleGAN. Then, we use DSPA to
%sample 400 stable diffusion-generated Van Gogh style images. Further,
%we use the same 6,287 content images and 800 Van Gogh style
%images to train CycleGAN-aug (400 human-created style images
%and 400 stable diffusion-generated style images). Similarly, To train
%CycleGAN-aug on Monet style, we utilize DSPA to expand Monet
%style images from 1072 to 2144. As shown in Fig.~\ref{image9}, the quality
%of the stylized images generated by CycleGAN-aug significantly
%improved. More qualitative and quantitative experimetns are shown in supplementary materials.
\begin{table*}[htb]
	\caption{The number of style images to train CycleGAN and CyleGAN-aug respectively. FID estimates the divergence between human-created artistic images and synthesized stylized images (lower is better).}
	\centering
	\setlength{\tabcolsep}{0.1cm}
	\begin{center}		
		\begin{tabular}{c|c|cccccccc}
			\hline  & Methods & Van Gogh & Morisot & Ukiyoe & Monet & Cezanne & Gauguin & Peploe&Rroerich \\
			\hline Number of  & CycleGAN & 400 & 1072 & 584 & 130 & 584
			& 293 & 204&859 \\
			style images& CycleGAN-aug & 800 & 2144 & 1168 & 260 & 1168 & 586 & 408&1718 \\
			\hline \multirow{2}{*}{FID $\downarrow$} & CycleGAN & 112.04 & 204.74 & 140.78 & 170.50 & 151.14 & 160.00 & 167.77&157.82 \\
			& CycleGAN-aug & \textbf{106.18} & \textbf{188.78} & \textbf{113.30} & \textbf{128.79} & \textbf{123.28} & \textbf{139.58} & \textbf{157.70}&\textbf{152.67} \\
			\hline
		\end{tabular}
	\end{center}
	\label{table2}
\end{table*}
\begin{figure*}[htb]	
	\centering
	\includegraphics[width=1.02\textwidth]{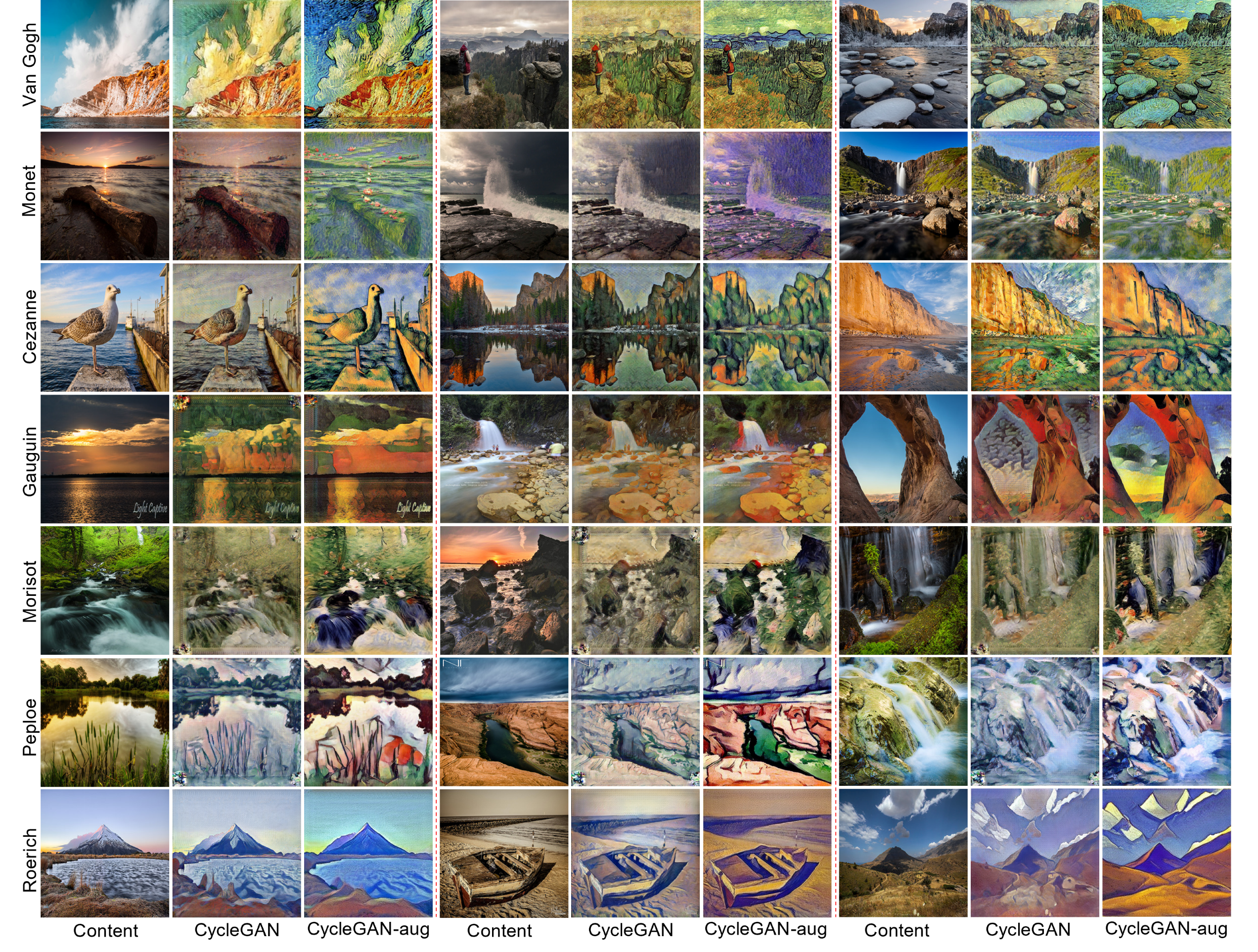} 
	\caption{Artistic stylized images generated by CycleGAN
		and CycleGAN-aug respectively.}
	\label{image10}
\end{figure*}

\textbf{Data Augmentation.}
To prove that DSPA can generate some data, which can help some existing methods are limited to the amount of data, we retrain the CycleGAN with the unaugmented data and CycleGAN-aug with the
augmented data (i.e., The augmented data includes human-created
style images and stable diffusion-generated style images). Specifically, we collect 6,287 content images from CycleGAN~\cite{zhu2017unpaired}.
Then, we collected some style images from WikiArt~\cite{nichol2016painter}. The size of each artwork collection is 400 for Van Gogh, 1072 for Monet, 584 for Cezanne, 130 for Morisot, 293 for Gauguin, 131 for Morisot, 204 for People and 859 for Roerich. We use these images as the unaugmented data. To further demonstrate the effectiveness of our method on data enhancement, we use DSPA to randomly
sample some style images, including 400 for Van Gogh, 1072 for Monet, 584 for Cezanne, 130 for Morisot, 293 for Gauguin, 131 for Morisot, 204 for People and 859 for Roerich. We merge these style images with the unaugmented data and retrain CycleGAN-aug. As shown in Tab.~\ref{table2}, we show the number of style images to train CycleGAN and CycleGAN-aug, respectively. It is worth noting that CycleGAN and CycleGAN-aug share the same network structure; we download code from  \url{https://github.com/junyanz/pytorch-CycleGAN-and-pix2pix}. To measure the image quality from a qualitative perspective, we utilize the
widely-used FID (Fréchet Inception Distance), which can estimate
the distribution divergence between human-created style images
and generated stylized images in a deep feature space~\cite{szegedy2016rethinking}. For CycleGAN and CycleGAN-aug, we randomly collect 6,000 content images from COCO~\cite{lin2014microsoft}, achieving
6,000 stylized images to compute FID between 6,000 stylized images
style images. As shown in Table.~\ref{table2}, CycleGAN-aug achieves a better FID score (lower is better). It implies that sampling some images as data augmenting via the proposed DSPA to train a small model is very efficient.

\begin{figure}[htb]	
	\centering
	\includegraphics[width=0.48\textwidth,height=0.30\textwidth]{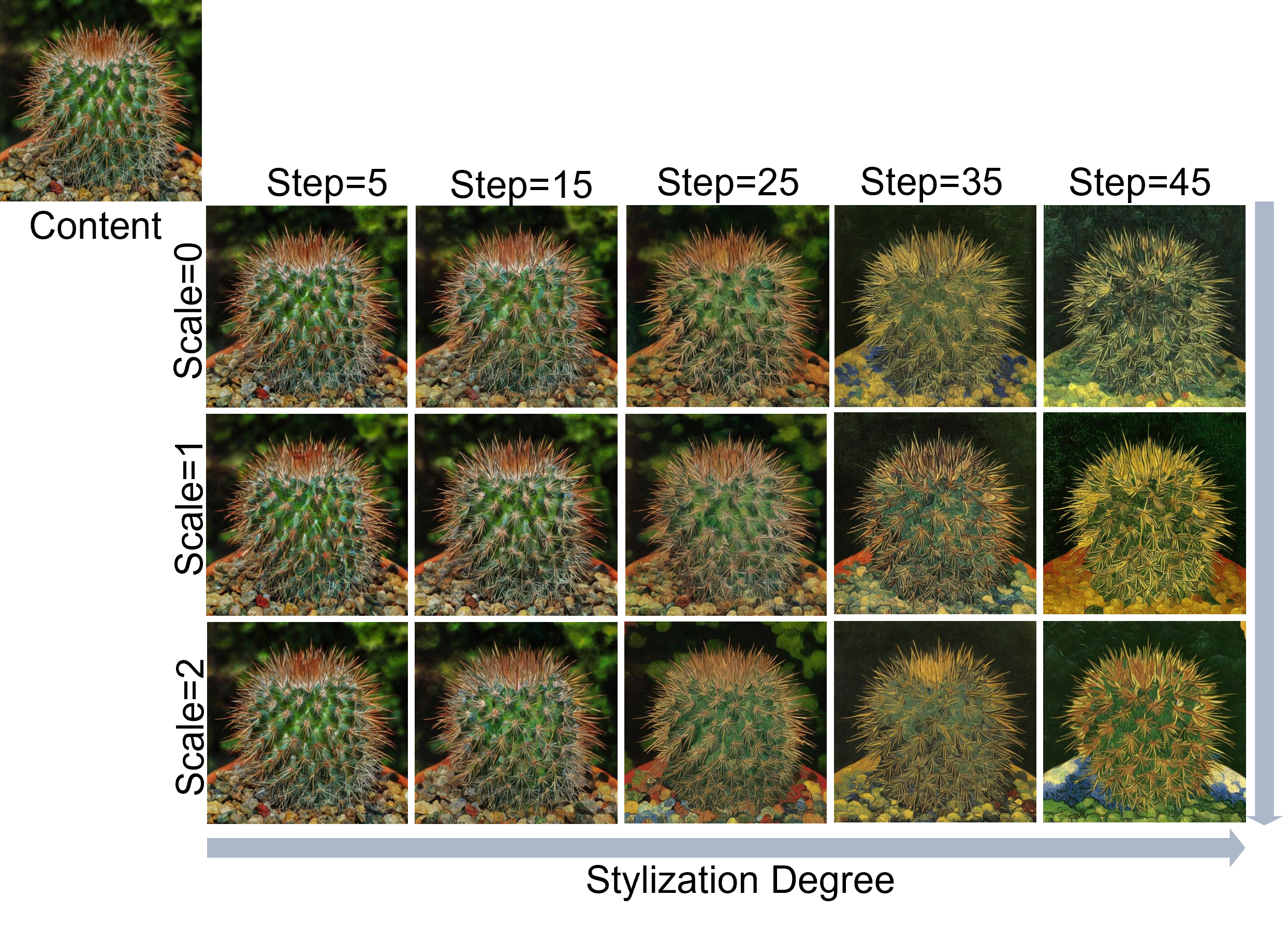} 
	\caption{Artistic stylized images (i.e., Van Gogh style) with
		scaling the variance of learnable parameters and denoising
		steps.}
	\label{image6}
\end{figure}

\textbf{Diversity and Stylization Degree Control}.
The optimization objective of DSPA is shown below:
\begin{equation}
\mathcal{L}_{d s p a}=\mathbb{E}_{\mathrm{x}, \omega, \epsilon, t}\left[\left\|\epsilon-\epsilon_\theta\left(\mathrm{x}_t, \boldsymbol{\mu}_{\mathrm{c}}+\omega \sigma_{\mathrm{c}}, t\right)\right\|_2^2\right]
\end{equation}
As shown in Fig.~\ref{image6}, in inference, we can multiply different scale factors $\gamma$ to
the variance $\sigma_{c}$ of Eq. 1 to control the diversity of stylized
images. The higher the value of $\gamma$, the greater the diversity. It is worth
noting that the scale factor $\gamma$ should be no more than 5. In addition, the denoising step is essential for the degree of
stylization. The more denoising steps, the more stylization.
%\begin{figure}[htb]	
%	\centering
%	\includegraphics[width=0.475\textwidth,height=0.25\textwidth]{image9.pdf} 
%	\caption{Artistic stylized images generated by CycleGAN
%		and CycleGAN-aug respectively. More qualitative and quantitative experimetns are shown in supplementary materials. }
%	\label{image9}
%\end{figure}

\begin{figure}[htb]	
	\centering
	\includegraphics[width=0.475\textwidth,height=0.21\textwidth]{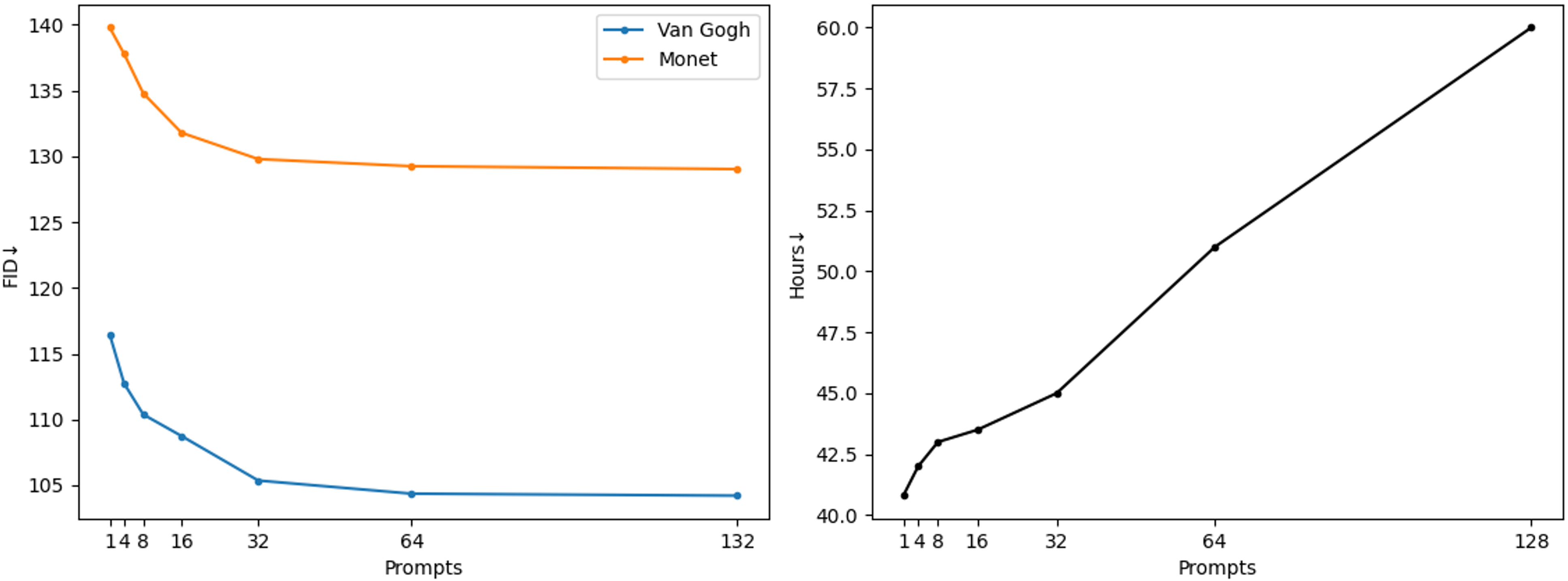} 
	\caption{With more prompts, the performance increases regarding FID, but computational cost also increases. However, when the number of learnable parameters reaches 32, there is no significant improvement in FID. This also strongly proves that compared with Artbank~\cite{zhang2024artbank}, using multiple learnable parameters is better than using a single learnable parameter.}
	\label{image11}
\end{figure}
\begin{figure}[htb]	
	\centering
	\includegraphics[width=0.47\textwidth,height=0.21\textwidth]{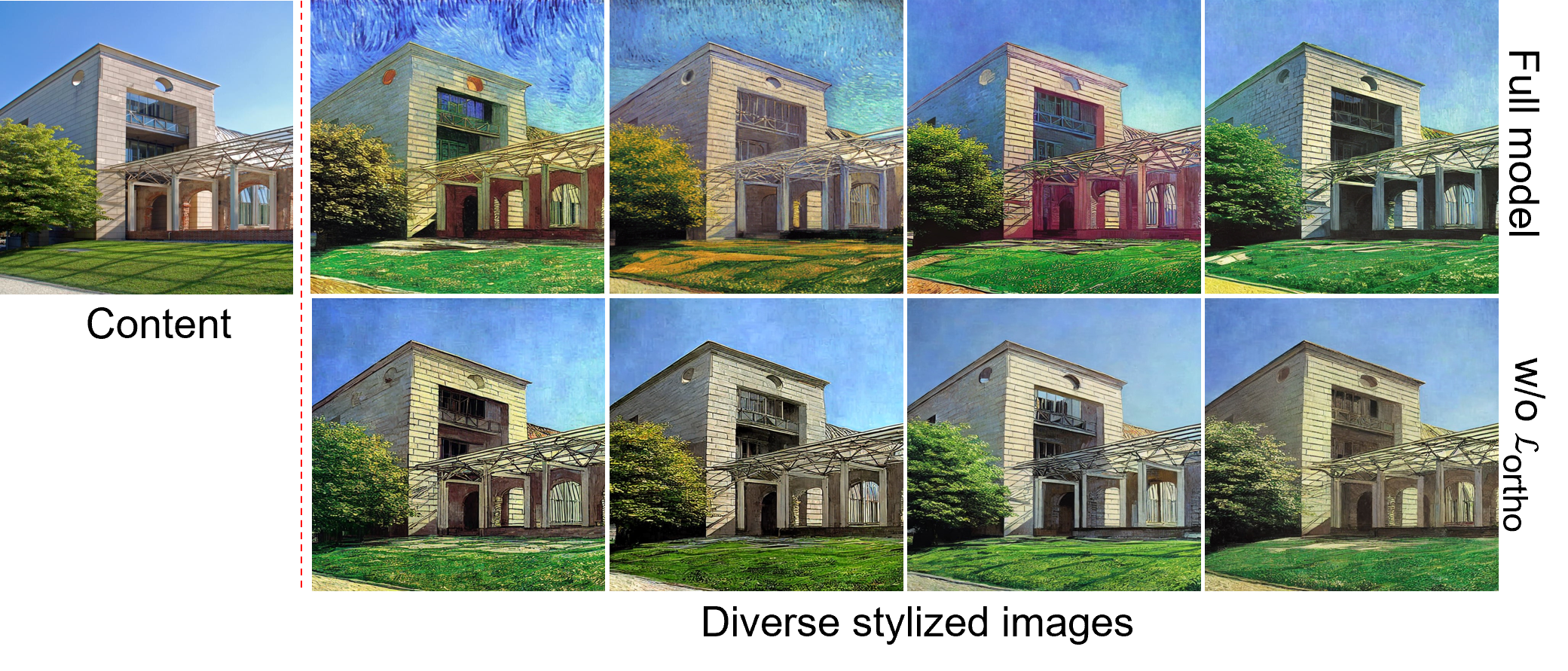} 
	\caption{The ablation study of $\mathcal{L}_{\text {ortho }}$. Please zoom-in for better comparison.}
	\label{image13}
\end{figure}
\textbf{Ablation Study of KCFP}. To verify that our proposed KCFP can better
preserve the content structure of the input content image, we compare it with stochastic inversion~\cite{zhang2024artbank} and
ControlNet~\cite{zhang2023adding} respectively. As shown in Fig.~\ref{image11}, if KCFP is removed, stylized images do not preserve the content structure of the input image. With stochastic inversion and ControlNet, stylized images have some difficulties in preserving the detailed content structure. With KCFP, stylized images can preserve the detailed content structure of the input image. This implies the superiority of our proposed KCFP. In addition, we calculate the average perceptual distance between
the input content image and the stylized image through LPIPS~\cite{choi2018stargan,zhu2017unpaired}.
Specifically, we randomly collect 500 input content images, generating
500 Van Gogh and Monet stylized images, respectively, with or without content preservation methods. As shown in Fig.~\ref{image11}, our proposed method achieves a
better LPIPS (lower is better).

\textbf{Ablation Study of DSPA}. In DSPA, the number of prompts is important in affecting the quality of stylized images. However, more prompts mean better image quality (lower FID) and more computational burden (more hours). In this experiment, we utilize two RTX 3090 GPUs to train DSPA. We show the training time (hours) in Fig.~\ref{image11}. As the number of prompts to train increases, so does the computational burden required. Considering stylized image quality and computational cost, we choose a prompt number of 32. As shown in Fig.~\ref{image13}, if $\mathcal{L}_{\text {ortho}}$ is removed, the stylized images show less diveristy.

%\subsection{Discussions and Limitations}
%Although our proposed DyArtbank can generate diverse and highly
%realistic artistic stylized images, the stylization examples show poor
%quality in some cases. In particular, the input inference images have some modern objects that do not exist in the artists’ artworks, such
%as airplanes, skyscrapers, etc.
\section{Limitation}
Although the method we proposed can generate diverse and highly realistic artistic stylized images, it is undeniable that our proposed DPSA module learns not only the style information from the collection of artworks but also the subjects (such as people and objects) present in that collection. However, the paintings of artists often exhibit a high degree of abstraction, and since artists do not depict modern objects (like buses and skyscrapers) in their works, this can lead to poor image quality during inference when encountering subjects that are absent from the artwork collection. Additionally, while our DSPA uses 32 learnable parameters, having more learnable parameters can enhance the quality of the generated images but also results in a polynomial increase in computational complexity.
\section{Conclusion}
In this paper, we introduce a novel framework called DyArtbank.
It can dig out the massive prior knowledge from the pre-trained stable diffusion and condition it to generate diverse and highly realistic
artistic stylized images. Besides, we propose a Dynamic Style
Prompt Artbank (DSPA), which effectively learns and stores the
style information from the collection of artworks and dynamically
conditions pre-trained stable diffusion to generate some diverse
artistic stylized images. DSPA can also support learning style
information from dozens or hundreds of artistic style images and utilize the learned style information to generate some artistic style
samples, which provides a new idea for data augmentation. Finally,
we propose a Key Content Feature Prompt (KCFP) module to provide
sufficient content prompts, guiding pre-trained SD to preserve
the content structure of input content images. To our knowledge,
our proposed approach is the first to use pre-trained stable diffusion
for diverse style transfer. In the future, we will focus on addressing the following two issues: 1) Although the diversity of stylized images can be controlled by applying different scale factors $\gamma$ to the variance $\sigma_{c}$ in Eq. 1, where larger scale factors can yield greater diversity, excessively large scale factors may render the guiding conditions ineffective. We aim to improve this aspect by designing more effective methods to achieve diverse results. 2) During inference, when the input image contains subjects that are not present in the training data, the quality of the generated images may suffer significantly.

\section{CRediT authorship contribution statement}
\textbf{Zhanjie Zhang:} Conceptualization, Methodology, Software, Writing – original draft.
\textbf{Quanwei Zhang:} Conceptualization, Methodology, Writing – original draft.
\textbf{Guangyuan Li:} Conceptualization, Methodology, Writing – original draft.
\textbf{Junsheng Luan:} Conceptualization, Methodology,
Writing – review editing. \textbf{Mengyuan Yang}: Software, Investigation, Data
curation, Validation, Writing – review editing. \textbf{Yun Wang:}
Software, Validation, Data curation, Writing – review editing.
\textbf{Lei Zhao:} Supervision, Writing – review editing.

\section{Declaration of Competing Interest}
The authors declare that they have no known competing financial interests or personal relationships that could have appeared to influence the work reported in this paper.

\section{Data availability}
Data will be made available on request.

\section{Acknowledgments}
This work was supported in part by Zhejiang Province Program
(2024C01110, 2022C01222, 2023C03199, 2023C03201), the National
Program of China (62172365, 2021YFF0900604, 19ZDA197), Macau
project: Key technology research and display system development for
new personalized controllable dressing dynamic display, Ningbo Science and Technology Plan Project (2022Z167, 2023Z137), and MOE
Frontier Science Center for Brain Science \& Brain-Machine Integration
(Zhejiang University).

\bibliographystyle{cas-model2-names}

\bibliography{cas-refs}

\begin{thebibliography}{53}
\expandafter\ifx\csname natexlab\endcsname\relax\def\natexlab#1{#1}\fi
\providecommand{\url}[1]{\texttt{#1}}
\providecommand{\href}[2]{#2}
\providecommand{\path}[1]{#1}
\providecommand{\DOIprefix}{doi:}
\providecommand{\ArXivprefix}{arXiv:}
\providecommand{\URLprefix}{URL: }
\providecommand{\Pubmedprefix}{pmid:}
\providecommand{\doi}[1]{\href{http://dx.doi.org/#1}{\path{#1}}}
\providecommand{\Pubmed}[1]{\href{pmid:#1}{\path{#1}}}
\providecommand{\bibinfo}[2]{#2}
\ifx\xfnm\relax \def\xfnm[#1]{\unskip,\space#1}\fi
%Type = Inproceedings
\bibitem[{Agustsson and Timofte(2017)}]{agustsson2017ntire}
\bibinfo{author}{Agustsson, E.}, \bibinfo{author}{Timofte, R.},
  \bibinfo{year}{2017}.
\newblock \bibinfo{title}{Ntire 2017 challenge on single image
  super-resolution: Dataset and study}, in: \bibinfo{booktitle}{Proceedings of
  the IEEE conference on computer vision and pattern recognition workshops},
  pp. \bibinfo{pages}{126--135}.
%Type = Article
\bibitem[{Chen et~al.(2024)Chen, Li, Ma, Qian, Wang and Li}]{chen2024towards}
\bibinfo{author}{Chen, B.}, \bibinfo{author}{Li, Q.}, \bibinfo{author}{Ma, R.},
  \bibinfo{author}{Qian, X.}, \bibinfo{author}{Wang, X.}, \bibinfo{author}{Li,
  X.}, \bibinfo{year}{2024}.
\newblock \bibinfo{title}{Towards the generalization of time series
  classification: A feature-level style transfer and multi-source transfer
  learning perspective}.
\newblock \bibinfo{journal}{Knowledge-Based Systems} , \bibinfo{pages}{112057}.
%Type = Inproceedings
\bibitem[{Chen et~al.(2021)Chen, Zhao, Zhang, Wang, Zuo, Li, Xing and
  Lu}]{chen2021diverse}
\bibinfo{author}{Chen, H.}, \bibinfo{author}{Zhao, L.}, \bibinfo{author}{Zhang,
  H.}, \bibinfo{author}{Wang, Z.}, \bibinfo{author}{Zuo, Z.},
  \bibinfo{author}{Li, A.}, \bibinfo{author}{Xing, W.}, \bibinfo{author}{Lu,
  D.}, \bibinfo{year}{2021}.
\newblock \bibinfo{title}{Diverse image style transfer via invertible
  cross-space mapping}, in: \bibinfo{booktitle}{2021 IEEE/CVF International
  Conference on Computer Vision (ICCV)}, \bibinfo{organization}{IEEE Computer
  Society}. pp. \bibinfo{pages}{14860--14869}.
%Type = Inproceedings
\bibitem[{Chen et~al.(2023)Chen, Ji, Zhang, Chu, Zuo, Zhao, Xing and
  Lu}]{chen2023testnerf}
\bibinfo{author}{Chen, J.}, \bibinfo{author}{Ji, B.}, \bibinfo{author}{Zhang,
  Z.}, \bibinfo{author}{Chu, T.}, \bibinfo{author}{Zuo, Z.},
  \bibinfo{author}{Zhao, L.}, \bibinfo{author}{Xing, W.}, \bibinfo{author}{Lu,
  D.}, \bibinfo{year}{2023}.
\newblock \bibinfo{title}{Testnerf: text-driven 3d style transfer via
  cross-modal learning}, in: \bibinfo{booktitle}{Proceedings of the
  Thirty-Second International Joint Conference on Artificial Intelligence}, pp.
  \bibinfo{pages}{5788--5796}.
%Type = Inproceedings
\bibitem[{Cheng et~al.(2023)Cheng, Wu, Jaiswal, Zhang, Natarajan and
  Natarajan}]{cheng2023user}
\bibinfo{author}{Cheng, J.}, \bibinfo{author}{Wu, Y.},
  \bibinfo{author}{Jaiswal, A.}, \bibinfo{author}{Zhang, X.},
  \bibinfo{author}{Natarajan, P.}, \bibinfo{author}{Natarajan, P.},
  \bibinfo{year}{2023}.
\newblock \bibinfo{title}{User-controllable arbitrary style transfer via
  entropy regularization}, in: \bibinfo{booktitle}{Proceedings of the AAAI
  Conference on Artificial Intelligence}, pp. \bibinfo{pages}{433--441}.
%Type = Inproceedings
\bibitem[{Choi et~al.(2018)Choi, Choi, Kim, Ha, Kim and Choo}]{choi2018stargan}
\bibinfo{author}{Choi, Y.}, \bibinfo{author}{Choi, M.}, \bibinfo{author}{Kim,
  M.}, \bibinfo{author}{Ha, J.W.}, \bibinfo{author}{Kim, S.},
  \bibinfo{author}{Choo, J.}, \bibinfo{year}{2018}.
\newblock \bibinfo{title}{Stargan: Unified generative adversarial networks for
  multi-domain image-to-image translation}, in: \bibinfo{booktitle}{Proceedings
  of the IEEE conference on computer vision and pattern recognition}, pp.
  \bibinfo{pages}{8789--8797}.
%Type = Inproceedings
\bibitem[{Chu et~al.(2024)Chu, Xing, Chen, Wang, Sun, Zhao, Chen and
  Lin}]{chu2024attack}
\bibinfo{author}{Chu, T.}, \bibinfo{author}{Xing, W.}, \bibinfo{author}{Chen,
  J.}, \bibinfo{author}{Wang, Z.}, \bibinfo{author}{Sun, J.},
  \bibinfo{author}{Zhao, L.}, \bibinfo{author}{Chen, H.}, \bibinfo{author}{Lin,
  H.}, \bibinfo{year}{2024}.
\newblock \bibinfo{title}{Attack deterministic conditional image generative
  models for diverse and controllable generation}, in:
  \bibinfo{booktitle}{Proceedings of the AAAI Conference on Artificial
  Intelligence}, pp. \bibinfo{pages}{1362--1370}.
%Type = Inproceedings
\bibitem[{Chung et~al.(2024)Chung, Hyun and Heo}]{chung2024style}
\bibinfo{author}{Chung, J.}, \bibinfo{author}{Hyun, S.}, \bibinfo{author}{Heo,
  J.P.}, \bibinfo{year}{2024}.
\newblock \bibinfo{title}{Style injection in diffusion: A training-free
  approach for adapting large-scale diffusion models for style transfer}, in:
  \bibinfo{booktitle}{Proceedings of the IEEE/CVF Conference on Computer Vision
  and Pattern Recognition}, pp. \bibinfo{pages}{8795--8805}.
%Type = Article
\bibitem[{Cuturi(2013)}]{cuturi2013sinkhorn}
\bibinfo{author}{Cuturi, M.}, \bibinfo{year}{2013}.
\newblock \bibinfo{title}{Sinkhorn distances: Lightspeed computation of optimal
  transport}.
\newblock \bibinfo{journal}{Advances in neural information processing systems}
  \bibinfo{volume}{26}.
%Type = Inproceedings
\bibitem[{Deng et~al.(2022)Deng, Tang, Dong, Ma, Pan, Wang and
  Xu}]{deng2022stytr2}
\bibinfo{author}{Deng, Y.}, \bibinfo{author}{Tang, F.}, \bibinfo{author}{Dong,
  W.}, \bibinfo{author}{Ma, C.}, \bibinfo{author}{Pan, X.},
  \bibinfo{author}{Wang, L.}, \bibinfo{author}{Xu, C.}, \bibinfo{year}{2022}.
\newblock \bibinfo{title}{Stytr2: Image style transfer with transformers}, in:
  \bibinfo{booktitle}{Proceedings of the IEEE/CVF conference on computer vision
  and pattern recognition}, pp. \bibinfo{pages}{11326--11336}.
%Type = Inproceedings
\bibitem[{Gal et~al.()Gal, Alaluf, Atzmon, Patashnik, Bermano, Chechik and
  Cohen-or}]{galimage}
\bibinfo{author}{Gal, R.}, \bibinfo{author}{Alaluf, Y.},
  \bibinfo{author}{Atzmon, Y.}, \bibinfo{author}{Patashnik, O.},
  \bibinfo{author}{Bermano, A.H.}, \bibinfo{author}{Chechik, G.},
  \bibinfo{author}{Cohen-or, D.}, .
\newblock \bibinfo{title}{An image is worth one word: Personalizing
  text-to-image generation using textual inversion}, in:
  \bibinfo{booktitle}{The Eleventh International Conference on Learning
  Representations}.
%Type = Article
\bibitem[{He et~al.(2024)He, Huang, Yuan, Zhong, Pun and
  Zeng}]{he2024progressive}
\bibinfo{author}{He, Z.}, \bibinfo{author}{Huang, G.}, \bibinfo{author}{Yuan,
  X.}, \bibinfo{author}{Zhong, G.}, \bibinfo{author}{Pun, C.M.},
  \bibinfo{author}{Zeng, Y.}, \bibinfo{year}{2024}.
\newblock \bibinfo{title}{Progressive normalizing flow with learnable spectrum
  transform for style transfer}.
\newblock \bibinfo{journal}{Knowledge-Based Systems} \bibinfo{volume}{284},
  \bibinfo{pages}{111277}.
%Type = Article
\bibitem[{Heusel et~al.(2017)Heusel, Ramsauer, Unterthiner, Nessler and
  Hochreiter}]{heusel2017gans}
\bibinfo{author}{Heusel, M.}, \bibinfo{author}{Ramsauer, H.},
  \bibinfo{author}{Unterthiner, T.}, \bibinfo{author}{Nessler, B.},
  \bibinfo{author}{Hochreiter, S.}, \bibinfo{year}{2017}.
\newblock \bibinfo{title}{Gans trained by a two time-scale update rule converge
  to a local nash equilibrium}.
\newblock \bibinfo{journal}{Advances in neural information processing systems}
  \bibinfo{volume}{30}.
%Type = Inproceedings
\bibitem[{Huang and Belongie(2017)}]{huang2017arbitrary}
\bibinfo{author}{Huang, X.}, \bibinfo{author}{Belongie, S.},
  \bibinfo{year}{2017}.
\newblock \bibinfo{title}{Arbitrary style transfer in real-time with adaptive
  instance normalization}, in: \bibinfo{booktitle}{Proceedings of the IEEE
  international conference on computer vision}, pp.
  \bibinfo{pages}{1501--1510}.
%Type = Inproceedings
\bibitem[{Huang et~al.(2018)Huang, Liu, Belongie and
  Kautz}]{huang2018multimodal}
\bibinfo{author}{Huang, X.}, \bibinfo{author}{Liu, M.Y.},
  \bibinfo{author}{Belongie, S.}, \bibinfo{author}{Kautz, J.},
  \bibinfo{year}{2018}.
\newblock \bibinfo{title}{Multimodal unsupervised image-to-image translation},
  in: \bibinfo{booktitle}{Proceedings of the European conference on computer
  vision (ECCV)}, pp. \bibinfo{pages}{172--189}.
%Type = Article
\bibitem[{Khosla et~al.(2020)Khosla, Teterwak, Wang, Sarna, Tian, Isola,
  Maschinot, Liu and Krishnan}]{khosla2020supervised}
\bibinfo{author}{Khosla, P.}, \bibinfo{author}{Teterwak, P.},
  \bibinfo{author}{Wang, C.}, \bibinfo{author}{Sarna, A.},
  \bibinfo{author}{Tian, Y.}, \bibinfo{author}{Isola, P.},
  \bibinfo{author}{Maschinot, A.}, \bibinfo{author}{Liu, C.},
  \bibinfo{author}{Krishnan, D.}, \bibinfo{year}{2020}.
\newblock \bibinfo{title}{Supervised contrastive learning}.
\newblock \bibinfo{journal}{Advances in neural information processing systems}
  \bibinfo{volume}{33}, \bibinfo{pages}{18661--18673}.
%Type = Article
\bibitem[{Kingma and Welling(2013)}]{kingma2013auto}
\bibinfo{author}{Kingma, D.P.}, \bibinfo{author}{Welling, M.},
  \bibinfo{year}{2013}.
\newblock \bibinfo{title}{Auto-encoding variational bayes}.
\newblock \bibinfo{journal}{arXiv preprint arXiv:1312.6114} .
%Type = Inproceedings
\bibitem[{Li et~al.(2024)Li, Rao, Mo, Zhang, Xing and Zhao}]{li2024rethinking}
\bibinfo{author}{Li, G.}, \bibinfo{author}{Rao, C.}, \bibinfo{author}{Mo, J.},
  \bibinfo{author}{Zhang, Z.}, \bibinfo{author}{Xing, W.},
  \bibinfo{author}{Zhao, L.}, \bibinfo{year}{2024}.
\newblock \bibinfo{title}{Rethinking diffusion model for multi-contrast mri
  super-resolution}, in: \bibinfo{booktitle}{Proceedings of the IEEE/CVF
  Conference on Computer Vision and Pattern Recognition}, pp.
  \bibinfo{pages}{11365--11374}.
%Type = Inproceedings
\bibitem[{Li et~al.(2023a)Li, Xing, Zhao, Lan, Sun, Zhang, Zhang, Lin and
  Lin}]{li2023self}
\bibinfo{author}{Li, G.}, \bibinfo{author}{Xing, W.}, \bibinfo{author}{Zhao,
  L.}, \bibinfo{author}{Lan, Z.}, \bibinfo{author}{Sun, J.},
  \bibinfo{author}{Zhang, Z.}, \bibinfo{author}{Zhang, Q.},
  \bibinfo{author}{Lin, H.}, \bibinfo{author}{Lin, Z.}, \bibinfo{year}{2023}a.
\newblock \bibinfo{title}{Self-reference image super-resolution via pre-trained
  diffusion large model and window adjustable transformer}, in:
  \bibinfo{booktitle}{Proceedings of the 31st ACM International Conference on
  Multimedia}, pp. \bibinfo{pages}{7981--7992}.
%Type = Inproceedings
\bibitem[{Li et~al.(2023b)Li, Zhao, Sun, Lan, Zhang, Chen, Lin, Lin and
  Xing}]{li2023rethinking}
\bibinfo{author}{Li, G.}, \bibinfo{author}{Zhao, L.}, \bibinfo{author}{Sun,
  J.}, \bibinfo{author}{Lan, Z.}, \bibinfo{author}{Zhang, Z.},
  \bibinfo{author}{Chen, J.}, \bibinfo{author}{Lin, Z.}, \bibinfo{author}{Lin,
  H.}, \bibinfo{author}{Xing, W.}, \bibinfo{year}{2023}b.
\newblock \bibinfo{title}{Rethinking multi-contrast mri super-resolution:
  Rectangle-window cross-attention transformer and arbitrary-scale upsampling},
  in: \bibinfo{booktitle}{Proceedings of the IEEE/CVF International Conference
  on Computer Vision}, pp. \bibinfo{pages}{21230--21240}.
%Type = Inproceedings
\bibitem[{Li et~al.(2022)Li, Li, Xiong and Hoi}]{li2022blip}
\bibinfo{author}{Li, J.}, \bibinfo{author}{Li, D.}, \bibinfo{author}{Xiong,
  C.}, \bibinfo{author}{Hoi, S.}, \bibinfo{year}{2022}.
\newblock \bibinfo{title}{Blip: Bootstrapping language-image pre-training for
  unified vision-language understanding and generation}, in:
  \bibinfo{booktitle}{International conference on machine learning},
  \bibinfo{organization}{PMLR}. pp. \bibinfo{pages}{12888--12900}.
%Type = Article
\bibitem[{Li et~al.(2023c)Li, Wu, Xu and Yao}]{li2023soft}
\bibinfo{author}{Li, J.}, \bibinfo{author}{Wu, L.}, \bibinfo{author}{Xu, D.},
  \bibinfo{author}{Yao, S.}, \bibinfo{year}{2023}c.
\newblock \bibinfo{title}{Soft multimodal style transfer via optimal
  transport}.
\newblock \bibinfo{journal}{Knowledge-Based Systems} \bibinfo{volume}{271},
  \bibinfo{pages}{110542}.
%Type = Inproceedings
\bibitem[{Lin et~al.(2014)Lin, Maire, Belongie, Hays, Perona, Ramanan,
  Doll{\'a}r and Zitnick}]{lin2014microsoft}
\bibinfo{author}{Lin, T.Y.}, \bibinfo{author}{Maire, M.},
  \bibinfo{author}{Belongie, S.}, \bibinfo{author}{Hays, J.},
  \bibinfo{author}{Perona, P.}, \bibinfo{author}{Ramanan, D.},
  \bibinfo{author}{Doll{\'a}r, P.}, \bibinfo{author}{Zitnick, C.L.},
  \bibinfo{year}{2014}.
\newblock \bibinfo{title}{Microsoft coco: Common objects in context}, in:
  \bibinfo{booktitle}{Computer Vision--ECCV 2014: 13th European Conference,
  Zurich, Switzerland, September 6-12, 2014, Proceedings, Part V 13},
  \bibinfo{organization}{Springer}. pp. \bibinfo{pages}{740--755}.
%Type = Inproceedings
\bibitem[{Liu et~al.(2024a)Liu, He, Lin and Wen}]{liu2024dual}
\bibinfo{author}{Liu, M.}, \bibinfo{author}{He, S.}, \bibinfo{author}{Lin, S.},
  \bibinfo{author}{Wen, B.}, \bibinfo{year}{2024}a.
\newblock \bibinfo{title}{Dual-head genre-instance transformer network for
  arbitrary style transfer}, in: \bibinfo{booktitle}{Proceedings of the 32nd
  ACM International Conference on Multimedia}, pp. \bibinfo{pages}{6024--6032}.
%Type = Article
\bibitem[{Liu et~al.(2024b)Liu, Lin, Zhang, Zha and Wen}]{liu2024intrinsic}
\bibinfo{author}{Liu, M.}, \bibinfo{author}{Lin, S.}, \bibinfo{author}{Zhang,
  H.}, \bibinfo{author}{Zha, Z.}, \bibinfo{author}{Wen, B.},
  \bibinfo{year}{2024}b.
\newblock \bibinfo{title}{Intrinsic-style distribution matching for arbitrary
  style transfer}.
\newblock \bibinfo{journal}{Knowledge-Based Systems} \bibinfo{volume}{296},
  \bibinfo{pages}{111898}.
%Type = Article
\bibitem[{Liu et~al.(2021a)Liu, Wang, Ji, Ge and Chen}]{liu2021person}
\bibinfo{author}{Liu, M.}, \bibinfo{author}{Wang, K.}, \bibinfo{author}{Ji,
  R.}, \bibinfo{author}{Ge, S.S.}, \bibinfo{author}{Chen, J.},
  \bibinfo{year}{2021}a.
\newblock \bibinfo{title}{Person image generation with attention-based
  injection network}.
\newblock \bibinfo{journal}{Neurocomputing} \bibinfo{volume}{460},
  \bibinfo{pages}{345--359}.
%Type = Article
\bibitem[{Liu et~al.(2021b)Liu, Wang, Ji, Ge and Chen}]{liu2021pose}
\bibinfo{author}{Liu, M.}, \bibinfo{author}{Wang, K.}, \bibinfo{author}{Ji,
  R.}, \bibinfo{author}{Ge, S.S.}, \bibinfo{author}{Chen, J.},
  \bibinfo{year}{2021}b.
\newblock \bibinfo{title}{Pose transfer generation with semantic parsing
  attention network for person re-identification}.
\newblock \bibinfo{journal}{Knowledge-Based Systems} \bibinfo{volume}{223},
  \bibinfo{pages}{107024}.
%Type = Inproceedings
\bibitem[{Lu et~al.(2022)Lu, Liu, Zhang, Liu and Tian}]{lu2022prompt}
\bibinfo{author}{Lu, Y.}, \bibinfo{author}{Liu, J.}, \bibinfo{author}{Zhang,
  Y.}, \bibinfo{author}{Liu, Y.}, \bibinfo{author}{Tian, X.},
  \bibinfo{year}{2022}.
\newblock \bibinfo{title}{Prompt distribution learning}, in:
  \bibinfo{booktitle}{Proceedings of the IEEE/CVF Conference on Computer Vision
  and Pattern Recognition}, pp. \bibinfo{pages}{5206--5215}.
%Type = Inproceedings
\bibitem[{Nichol and Dhariwal(2021)}]{nichol2021improved}
\bibinfo{author}{Nichol, A.Q.}, \bibinfo{author}{Dhariwal, P.},
  \bibinfo{year}{2021}.
\newblock \bibinfo{title}{Improved denoising diffusion probabilistic models},
  in: \bibinfo{booktitle}{International conference on machine learning},
  \bibinfo{organization}{PMLR}. pp. \bibinfo{pages}{8162--8171}.
%Type = Article
\bibitem[{Nichol(2016)}]{nichol2016painter}
\bibinfo{author}{Nichol, K.}, \bibinfo{year}{2016}.
\newblock \bibinfo{title}{Painter by numbers, wikiart}.
\newblock \bibinfo{journal}{Kiri Nichol} \bibinfo{volume}{8}.
%Type = Inproceedings
\bibitem[{Park et~al.(2020)Park, Efros, Zhang and Zhu}]{park2020contrastive}
\bibinfo{author}{Park, T.}, \bibinfo{author}{Efros, A.A.},
  \bibinfo{author}{Zhang, R.}, \bibinfo{author}{Zhu, J.Y.},
  \bibinfo{year}{2020}.
\newblock \bibinfo{title}{Contrastive learning for unpaired image-to-image
  translation}, in: \bibinfo{booktitle}{Computer Vision--ECCV 2020: 16th
  European Conference, Glasgow, UK, August 23--28, 2020, Proceedings, Part IX
  16}, \bibinfo{organization}{Springer}. pp. \bibinfo{pages}{319--345}.
%Type = Article
\bibitem[{Podell et~al.(2023)Podell, English, Lacey, Blattmann, Dockhorn,
  M{\"u}ller, Penna and Rombach}]{podell2023sdxl}
\bibinfo{author}{Podell, D.}, \bibinfo{author}{English, Z.},
  \bibinfo{author}{Lacey, K.}, \bibinfo{author}{Blattmann, A.},
  \bibinfo{author}{Dockhorn, T.}, \bibinfo{author}{M{\"u}ller, J.},
  \bibinfo{author}{Penna, J.}, \bibinfo{author}{Rombach, R.},
  \bibinfo{year}{2023}.
\newblock \bibinfo{title}{Sdxl: Improving latent diffusion models for
  high-resolution image synthesis}.
\newblock \bibinfo{journal}{arXiv preprint arXiv:2307.01952} .
%Type = Article
\bibitem[{Qu et~al.(2024)Qu, Liu, Zhu, Nie and Zhang}]{qu2024source}
\bibinfo{author}{Qu, X.}, \bibinfo{author}{Liu, L.}, \bibinfo{author}{Zhu, L.},
  \bibinfo{author}{Nie, L.}, \bibinfo{author}{Zhang, H.}, \bibinfo{year}{2024}.
\newblock \bibinfo{title}{Source-free style-diversity adversarial domain
  adaptation with privacy-preservation for person re-identification}.
\newblock \bibinfo{journal}{Knowledge-Based Systems} \bibinfo{volume}{283},
  \bibinfo{pages}{111150}.
%Type = Inproceedings
\bibitem[{Radford et~al.(2021)Radford, Kim, Hallacy, Ramesh, Goh, Agarwal,
  Sastry, Askell, Mishkin, Clark et~al.}]{radford2021learning}
\bibinfo{author}{Radford, A.}, \bibinfo{author}{Kim, J.W.},
  \bibinfo{author}{Hallacy, C.}, \bibinfo{author}{Ramesh, A.},
  \bibinfo{author}{Goh, G.}, \bibinfo{author}{Agarwal, S.},
  \bibinfo{author}{Sastry, G.}, \bibinfo{author}{Askell, A.},
  \bibinfo{author}{Mishkin, P.}, \bibinfo{author}{Clark, J.}, et~al.,
  \bibinfo{year}{2021}.
\newblock \bibinfo{title}{Learning transferable visual models from natural
  language supervision}, in: \bibinfo{booktitle}{International conference on
  machine learning}, \bibinfo{organization}{PMLR}. pp.
  \bibinfo{pages}{8748--8763}.
%Type = Inproceedings
\bibitem[{Rombach et~al.(2022)Rombach, Blattmann, Lorenz, Esser and
  Ommer}]{rombach2022high}
\bibinfo{author}{Rombach, R.}, \bibinfo{author}{Blattmann, A.},
  \bibinfo{author}{Lorenz, D.}, \bibinfo{author}{Esser, P.},
  \bibinfo{author}{Ommer, B.}, \bibinfo{year}{2022}.
\newblock \bibinfo{title}{High-resolution image synthesis with latent diffusion
  models}, in: \bibinfo{booktitle}{Proceedings of the IEEE/CVF conference on
  computer vision and pattern recognition}, pp. \bibinfo{pages}{10684--10695}.
%Type = Inproceedings
\bibitem[{Sanakoyeu et~al.(2018)Sanakoyeu, Kotovenko, Lang and
  Ommer}]{sanakoyeu2018style}
\bibinfo{author}{Sanakoyeu, A.}, \bibinfo{author}{Kotovenko, D.},
  \bibinfo{author}{Lang, S.}, \bibinfo{author}{Ommer, B.},
  \bibinfo{year}{2018}.
\newblock \bibinfo{title}{A style-aware content loss for real-time hd style
  transfer}, in: \bibinfo{booktitle}{proceedings of the European conference on
  computer vision (ECCV)}, pp. \bibinfo{pages}{698--714}.
%Type = Inproceedings
\bibitem[{Sun et~al.(2024)Sun, Jiao, Li, Zhang, Zhao and
  Xing}]{sun20243dgstream}
\bibinfo{author}{Sun, J.}, \bibinfo{author}{Jiao, H.}, \bibinfo{author}{Li,
  G.}, \bibinfo{author}{Zhang, Z.}, \bibinfo{author}{Zhao, L.},
  \bibinfo{author}{Xing, W.}, \bibinfo{year}{2024}.
\newblock \bibinfo{title}{3dgstream: On-the-fly training of 3d gaussians for
  efficient streaming of photo-realistic free-viewpoint videos}, in:
  \bibinfo{booktitle}{Proceedings of the IEEE/CVF Conference on Computer Vision
  and Pattern Recognition}, pp. \bibinfo{pages}{20675--20685}.
%Type = Article
\bibitem[{Sun et~al.(2023)Sun, Zhang, Chen, Li, Ji, Zhao, Xing and
  Lin}]{sun2023vgos}
\bibinfo{author}{Sun, J.}, \bibinfo{author}{Zhang, Z.}, \bibinfo{author}{Chen,
  J.}, \bibinfo{author}{Li, G.}, \bibinfo{author}{Ji, B.},
  \bibinfo{author}{Zhao, L.}, \bibinfo{author}{Xing, W.}, \bibinfo{author}{Lin,
  H.}, \bibinfo{year}{2023}.
\newblock \bibinfo{title}{Vgos: Voxel grid optimization for view synthesis from
  sparse inputs}.
\newblock \bibinfo{journal}{arXiv preprint arXiv:2304.13386} .
%Type = Inproceedings
\bibitem[{Szegedy et~al.(2016)Szegedy, Vanhoucke, Ioffe, Shlens and
  Wojna}]{szegedy2016rethinking}
\bibinfo{author}{Szegedy, C.}, \bibinfo{author}{Vanhoucke, V.},
  \bibinfo{author}{Ioffe, S.}, \bibinfo{author}{Shlens, J.},
  \bibinfo{author}{Wojna, Z.}, \bibinfo{year}{2016}.
\newblock \bibinfo{title}{Rethinking the inception architecture for computer
  vision}, in: \bibinfo{booktitle}{Proceedings of the IEEE conference on
  computer vision and pattern recognition}, pp. \bibinfo{pages}{2818--2826}.
%Type = Article
\bibitem[{Wang et~al.(2024)Wang, Wang, Li, Zhang, Wu and Guo}]{wang2024cost}
\bibinfo{author}{Wang, Y.}, \bibinfo{author}{Wang, L.}, \bibinfo{author}{Li,
  K.}, \bibinfo{author}{Zhang, Y.}, \bibinfo{author}{Wu, D.O.},
  \bibinfo{author}{Guo, Y.}, \bibinfo{year}{2024}.
\newblock \bibinfo{title}{Cost volume aggregation in stereo matching revisited:
  A disparity classification perspective}.
\newblock \bibinfo{journal}{IEEE Transactions on Image Processing} .
%Type = Article
\bibitem[{Wang et~al.(2022a)Wang, Wang, Wang and Guo}]{wang2022spnet}
\bibinfo{author}{Wang, Y.}, \bibinfo{author}{Wang, L.}, \bibinfo{author}{Wang,
  H.}, \bibinfo{author}{Guo, Y.}, \bibinfo{year}{2022}a.
\newblock \bibinfo{title}{Spnet: Learning stereo matching with slanted plane
  aggregation}.
\newblock \bibinfo{journal}{IEEE Robotics and Automation Letters}
  \bibinfo{volume}{7}, \bibinfo{pages}{6258--6265}.
%Type = Inproceedings
\bibitem[{Wang et~al.(2022b)Wang, Zhang, Zhao, Zuo, Li, Xing and
  Lu}]{wang2022aesust}
\bibinfo{author}{Wang, Z.}, \bibinfo{author}{Zhang, Z.}, \bibinfo{author}{Zhao,
  L.}, \bibinfo{author}{Zuo, Z.}, \bibinfo{author}{Li, A.},
  \bibinfo{author}{Xing, W.}, \bibinfo{author}{Lu, D.}, \bibinfo{year}{2022}b.
\newblock \bibinfo{title}{Aesust: towards aesthetic-enhanced universal style
  transfer}, in: \bibinfo{booktitle}{Proceedings of the 30th ACM International
  Conference on Multimedia}, pp. \bibinfo{pages}{1095--1106}.
%Type = Inproceedings
\bibitem[{Wang et~al.(2020)Wang, Zhao, Chen, Qiu, Mo, Lin, Xing and
  Lu}]{wang2020diversified}
\bibinfo{author}{Wang, Z.}, \bibinfo{author}{Zhao, L.}, \bibinfo{author}{Chen,
  H.}, \bibinfo{author}{Qiu, L.}, \bibinfo{author}{Mo, Q.},
  \bibinfo{author}{Lin, S.}, \bibinfo{author}{Xing, W.}, \bibinfo{author}{Lu,
  D.}, \bibinfo{year}{2020}.
\newblock \bibinfo{title}{Diversified arbitrary style transfer via deep feature
  perturbation}, in: \bibinfo{booktitle}{Proceedings of the IEEE/CVF Conference
  on Computer Vision and Pattern Recognition}, pp. \bibinfo{pages}{7789--7798}.
%Type = Article
\bibitem[{Yang et~al.(2022)Yang, Chen, Zhang, Zhao and Lin}]{yang2022gating}
\bibinfo{author}{Yang, F.}, \bibinfo{author}{Chen, H.}, \bibinfo{author}{Zhang,
  Z.}, \bibinfo{author}{Zhao, L.}, \bibinfo{author}{Lin, H.},
  \bibinfo{year}{2022}.
\newblock \bibinfo{title}{Gating patternpyramid for diversified image style
  transfer}.
\newblock \bibinfo{journal}{Journal of Electronic Imaging}
  \bibinfo{volume}{31}, \bibinfo{pages}{063007--063007}.
%Type = Inproceedings
\bibitem[{Zhang et~al.(2023a)Zhang, Rao and Agrawala}]{zhang2023adding}
\bibinfo{author}{Zhang, L.}, \bibinfo{author}{Rao, A.},
  \bibinfo{author}{Agrawala, M.}, \bibinfo{year}{2023}a.
\newblock \bibinfo{title}{Adding conditional control to text-to-image diffusion
  models}, in: \bibinfo{booktitle}{Proceedings of the IEEE/CVF International
  Conference on Computer Vision}, pp. \bibinfo{pages}{3836--3847}.
%Type = Article
\bibitem[{Zhang et~al.(2021)Zhang, Zhang, Jia, He and
  Yang}]{zhang2021generating}
\bibinfo{author}{Zhang, T.}, \bibinfo{author}{Zhang, Z.}, \bibinfo{author}{Jia,
  W.}, \bibinfo{author}{He, X.}, \bibinfo{author}{Yang, J.},
  \bibinfo{year}{2021}.
\newblock \bibinfo{title}{Generating cartoon images from face photos with
  cycle-consistent adversarial networks}.
\newblock \bibinfo{journal}{Computers, Materials and Continua} .
%Type = Article
\bibitem[{Zhang et~al.(2023b)Zhang, Dong, Tang, Huang, Huang, Ma, Lee, Deussen
  and Xu}]{zhang2023prospect}
\bibinfo{author}{Zhang, Y.}, \bibinfo{author}{Dong, W.}, \bibinfo{author}{Tang,
  F.}, \bibinfo{author}{Huang, N.}, \bibinfo{author}{Huang, H.},
  \bibinfo{author}{Ma, C.}, \bibinfo{author}{Lee, T.Y.},
  \bibinfo{author}{Deussen, O.}, \bibinfo{author}{Xu, C.},
  \bibinfo{year}{2023}b.
\newblock \bibinfo{title}{Prospect: Prompt spectrum for attribute-aware
  personalization of diffusion models}.
\newblock \bibinfo{journal}{ACM Transactions on Graphics (TOG)}
  \bibinfo{volume}{42}, \bibinfo{pages}{1--14}.
%Type = Inproceedings
\bibitem[{Zhang et~al.(2023c)Zhang, Huang, Tang, Huang, Ma, Dong and
  Xu}]{zhang2023inversion}
\bibinfo{author}{Zhang, Y.}, \bibinfo{author}{Huang, N.},
  \bibinfo{author}{Tang, F.}, \bibinfo{author}{Huang, H.}, \bibinfo{author}{Ma,
  C.}, \bibinfo{author}{Dong, W.}, \bibinfo{author}{Xu, C.},
  \bibinfo{year}{2023}c.
\newblock \bibinfo{title}{Inversion-based style transfer with diffusion
  models}, in: \bibinfo{booktitle}{Proceedings of the IEEE/CVF conference on
  computer vision and pattern recognition}, pp. \bibinfo{pages}{10146--10156}.
%Type = Article
\bibitem[{Zhang et~al.(2023d)Zhang, Sun, Chen, Zhao, Ji, Lan, Li, Xing and
  Xu}]{zhang2023caster}
\bibinfo{author}{Zhang, Z.}, \bibinfo{author}{Sun, J.}, \bibinfo{author}{Chen,
  J.}, \bibinfo{author}{Zhao, L.}, \bibinfo{author}{Ji, B.},
  \bibinfo{author}{Lan, Z.}, \bibinfo{author}{Li, G.}, \bibinfo{author}{Xing,
  W.}, \bibinfo{author}{Xu, D.}, \bibinfo{year}{2023}d.
\newblock \bibinfo{title}{Caster: Cartoon style transfer via dynamic cartoon
  style casting}.
\newblock \bibinfo{journal}{Neurocomputing} \bibinfo{volume}{556},
  \bibinfo{pages}{126654}.
%Type = Article
\bibitem[{Zhang et~al.(2024a)Zhang, Sun, Li, Zhao, Zhang, Lan, Yin, Xing, Lin
  and Zuo}]{zhang2024rethink}
\bibinfo{author}{Zhang, Z.}, \bibinfo{author}{Sun, J.}, \bibinfo{author}{Li,
  G.}, \bibinfo{author}{Zhao, L.}, \bibinfo{author}{Zhang, Q.},
  \bibinfo{author}{Lan, Z.}, \bibinfo{author}{Yin, H.}, \bibinfo{author}{Xing,
  W.}, \bibinfo{author}{Lin, H.}, \bibinfo{author}{Zuo, Z.},
  \bibinfo{year}{2024}a.
\newblock \bibinfo{title}{Rethink arbitrary style transfer with transformer and
  contrastive learning}.
\newblock \bibinfo{journal}{Computer Vision and Image Understanding} ,
  \bibinfo{pages}{103951}.
%Type = Article
\bibitem[{Zhang et~al.(2024b)Zhang, Zhang, Lin, Xing, Mo, Huang, Xie, Li, Luan,
  Zhao et~al.}]{zhang2024towards}
\bibinfo{author}{Zhang, Z.}, \bibinfo{author}{Zhang, Q.}, \bibinfo{author}{Lin,
  H.}, \bibinfo{author}{Xing, W.}, \bibinfo{author}{Mo, J.},
  \bibinfo{author}{Huang, S.}, \bibinfo{author}{Xie, J.}, \bibinfo{author}{Li,
  G.}, \bibinfo{author}{Luan, J.}, \bibinfo{author}{Zhao, L.}, et~al.,
  \bibinfo{year}{2024}b.
\newblock \bibinfo{title}{Towards highly realistic artistic style transfer via
  stable diffusion with step-aware and layer-aware prompt}.
\newblock \bibinfo{journal}{The 33rd International Joint Conference on
  Artificial Intelligence} .
%Type = Inproceedings
\bibitem[{Zhang et~al.(2024c)Zhang, Zhang, Xing, Li, Zhao, Sun, Lan, Luan,
  Huang and Lin}]{zhang2024artbank}
\bibinfo{author}{Zhang, Z.}, \bibinfo{author}{Zhang, Q.},
  \bibinfo{author}{Xing, W.}, \bibinfo{author}{Li, G.}, \bibinfo{author}{Zhao,
  L.}, \bibinfo{author}{Sun, J.}, \bibinfo{author}{Lan, Z.},
  \bibinfo{author}{Luan, J.}, \bibinfo{author}{Huang, Y.},
  \bibinfo{author}{Lin, H.}, \bibinfo{year}{2024}c.
\newblock \bibinfo{title}{Artbank: Artistic style transfer with pre-trained
  diffusion model and implicit style prompt bank}, in:
  \bibinfo{booktitle}{Proceedings of the AAAI Conference on Artificial
  Intelligence}, pp. \bibinfo{pages}{7396--7404}.
%Type = Inproceedings
\bibitem[{Zhu et~al.(2017)Zhu, Park, Isola and Efros}]{zhu2017unpaired}
\bibinfo{author}{Zhu, J.Y.}, \bibinfo{author}{Park, T.},
  \bibinfo{author}{Isola, P.}, \bibinfo{author}{Efros, A.A.},
  \bibinfo{year}{2017}.
\newblock \bibinfo{title}{Unpaired image-to-image translation using
  cycle-consistent adversarial networks}, in: \bibinfo{booktitle}{Proceedings
  of the IEEE international conference on computer vision}, pp.
  \bibinfo{pages}{2223--2232}.

\end{thebibliography}


\end{document}